\documentclass{article} 
\usepackage{iclr2027_conference,times}

\usepackage{amsmath,amsfonts,bm}

\def\eqref#1{equation~\ref{#1}}

\def\1{\bm{1}}

\DeclareMathAlphabet{\mathsfit}{\encodingdefault}{\sfdefault}{m}{sl}
\SetMathAlphabet{\mathsfit}{bold}{\encodingdefault}{\sfdefault}{bx}{n}

\usepackage{hyperref}
\usepackage{url}
\usepackage{graphicx}
\usepackage{booktabs}
\usepackage{multirow}
\usepackage{makecell}
\usepackage{placeins}
\usepackage{bbm}
\usepackage{subcaption}
\usepackage{tikz}
\usepackage{pgfplots}
\usepackage{todonotes}
\usepackage{xcolor}
\usepackage{caption}
\usepackage{amsmath}
\usepackage{amssymb}
\usepackage{booktabs}
\usepackage{multirow}
\usepackage{amsmath}
\usepackage{natbib}
\usepackage{float}
\usepackage{booktabs}
\usepackage{graphicx}
\usepackage{wrapfig}
\usepackage{etoolbox}
\usepackage{array}
\usepackage{tabularx}
\usepackage{longtable}
\usepackage[most]{tcolorbox}

\title{WAM-Sim0.1: \underline{W}orld \underline{A}nchored \underline{M}emory Meets Online RL for Consistent and Realistic Driving \underline{Sim}ulation}

\author{
\textbf{Jiahao Cui}\textsuperscript{1,*},
\textbf{Jiaye Li}\textsuperscript{1,*},
\textbf{Wenhao Zhang}\textsuperscript{1,*},
\textbf{Yaoyi Li}\textsuperscript{2,*},
\textbf{Shan Luan}\textsuperscript{1,*},
\textbf{Quanhui Tang}\textsuperscript{1} \\
\textbf{Jianhua Han}\textsuperscript{2},
\textbf{Lin Cheng}\textsuperscript{2},
\textbf{Hang Xu}\textsuperscript{2},
\textbf{Siyu Zhu}\textsuperscript{1,\textdagger} \\
\textsuperscript{1}Fudan University \\
\textsuperscript{2}Shenzhen Yinwang Intelligent Technology Co., Ltd \\
\textsuperscript{*}Equal contribution. \textsuperscript{\textdagger}Corresponding author.
}

\iclrfinalcopy
\begin{document}

\maketitle
\lhead{}

\newcommand{\tablefont}{\scriptsize}
\AtBeginEnvironment{tabular}{\tablefont}
\AtBeginEnvironment{tabular*}{\tablefont}
\AtBeginEnvironment{tabularx}{\tablefont}

\begin{abstract}
Autoregressive driving world models systematically suffer from structural drift, identity drift, and quality decay over long rollout horizons due to transient conditioning signals that lack persistent state. 
We present WAM-Sim, a driving world model anchored by a persistent hierarchical memory and optimized via online flow-matching RL. 
WAM-Sim structures persistent context across three temporal and semantic scales: 
pose-aligned structural tokens encoding persistent road topology, 
read-only scene tokens supplying geolocated background appearance, 
and write-once instance tokens preserving individual traffic participant identities across occlusions and chunk boundaries. 
To restore visual sharpness and physical plausibility degraded by multi-scale conditioning, 
we introduce an online post-training framework that optimizes self-generated rollouts via flow-matching RL guided by hierarchy-aligned visual rewards and a vision-language model physical judge. 
The invariant memory empirically confines policy optimization within the anchored context, helping suppress semantic collapse and reward hacking,
while online RL systematically enforces physical commonsense and fine-grained visual fidelity. 
On the nuScenes benchmark, WAM-Sim achieves a superior generation fidelity (FID=5.75, FVD=32.40) and strong downstream perception performance (22.30 mAP, 70.32 road mIoU). 
On minute-scale rollouts of self-collected data, 
WAM-Sim substantially mitigates structural and identity drift, 
preserving high scene similarity and subject identity coherence across extended horizons.
\end{abstract}

\section{Introduction}
\label{sec:intro}

Video world models offer a scalable paradigm for controllable closed-loop simulation in autonomous driving, 
synthesizing driving observations conditioned on ego trajectories and agent controls~\citep{hu2023gaia,gao2024vista,wen2024panacea,russell2025gaia}. 
Despite recent progress, standard training and evaluation regimes remain predominantly constrained to short temporal clips and instantaneous point-estimate metrics. 
As rollouts extend toward minute-scale horizons, 
generation systematically degrades through three recurring failure modes: 
1) structural drift, where road geometry and background appearance detach from geographic reality; 
2) identity drift, where traffic participants undergo unconstrained appearance, category, or tracking mutations; 
and 3) quality decay, where accumulated autoregressive errors recursively compound over time. 
Although recent architectures incorporate extended context windows, temporal filtering, or instance masking~\citep{guo2024infinitydrive,huang2025selfforcing,liu2026rollingforcing,cheng2026m4world,basant2026omnidreams}, 
these capabilities operate in isolation without a persistent and addressable state.
We argue that long-horizon generation failure is not merely a function of limited context window length, 
but fundamentally demands persistent, multi-scale memory semantics to anchor identity, appearance and geometry.

In conventional driving world models, conditioning signals such as high-definition (HD) maps, 3D bounding boxes, and reference frames are provided as transient per-frame or per-chunk inputs. 
While transient conditioning constrains immediate frame synthesis, 
it fails to establish a persistent memory state across extended rollouts. 
Recent retrieval-augmented models and object-centric attention mechanisms demonstrate the value of persistent scene context and isolated participant conditioning~\citep{wu2025longtermspatialmemory,xiao2025worldmem,li2025vmem,jia2026spatialretrieval,seo2026seoulworldmodel,yang2026consisdrive}. 
However, existing methods deploy these mechanisms in isolation, 
failing to jointly preserve global route geometry, geolocated scene appearance, and individual object identities under dynamic mid-rollout interventions. 
Consequently, once generation extends beyond the initial context window,
persistent states must be reconstructed from noisy self-generated histories,
causing errors to compound over time.
Overcoming this bottleneck requires a unified memory architecture that
represents road structure, environmental appearance, and participant identity
as distinct visual tokens, maintains each according to its temporal
persistence, and retrieves each through mechanisms matched to its
spatio-temporal validity domain.

\begin{figure}[!t]
  \centering
  \includegraphics[width=\linewidth]{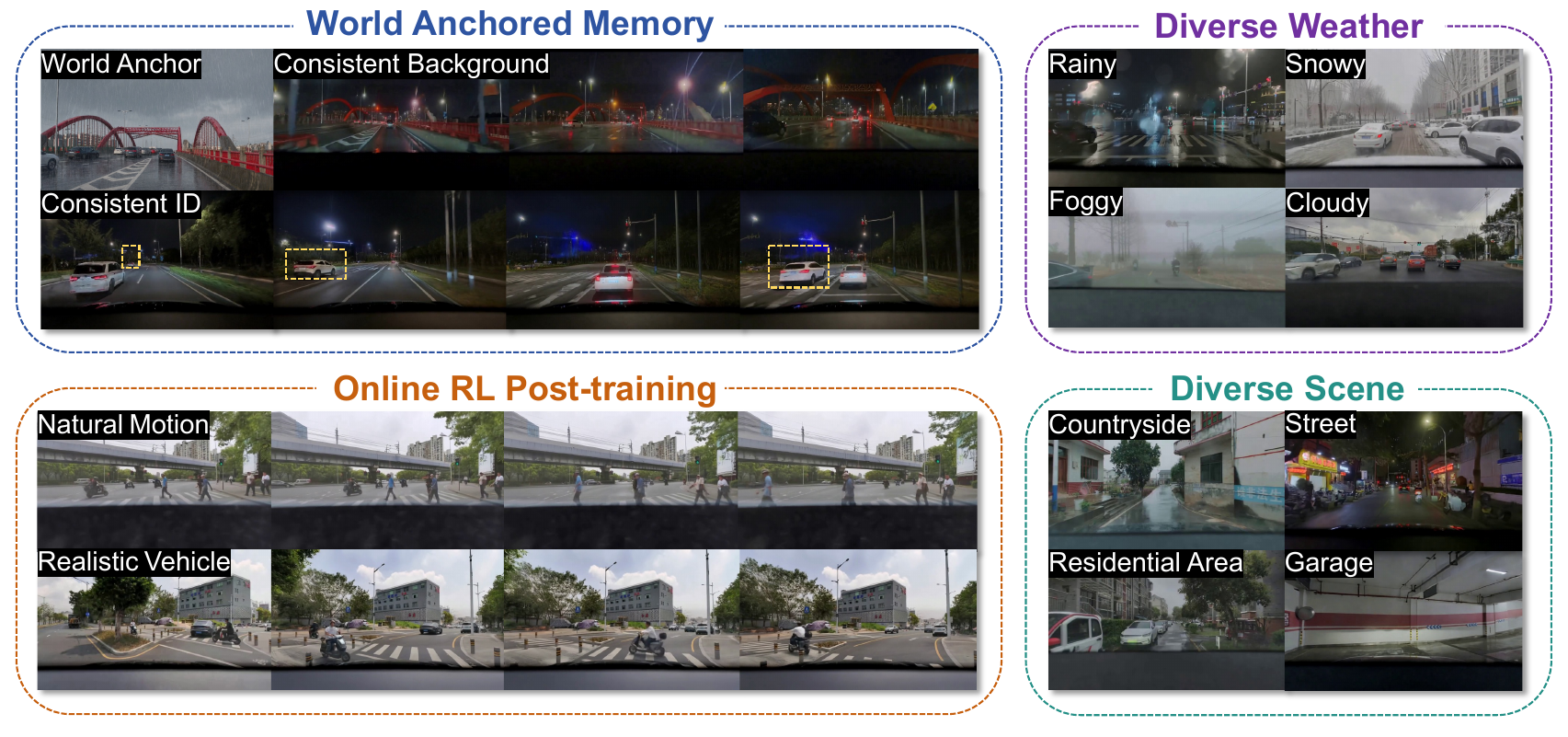}
  \caption{Overview of core capabilities in WAM-Sim.
  World anchored memory (top-left) anchors long-horizon rollouts, maintaining background consistency and persistent instance identities (e.g., tracked pedestrians, motor vehicles, and non-motorized vehicles) across extended horizons. 
  Online RL Post-training (bottom-left) restores visual fidelity and enforces physical commonsense, synthesizing natural agent dynamics (e.g., pedestrian gait) and realistic vehicle motion. 
  Furthermore, WAM-Sim generalizes across diverse weather conditions and environmental scenes (right).}
  \label{fig:ultra_long_horizon}
\end{figure}

To address this challenge, 
we present a driving world model anchored by a persistent world anchored memory. 
The memory organizes the driving world across three temporal and semantic scales: 
static structural tokens encoded from HD lane maps serve as persistent geometric anchors; 
read-only scene tokens retrieved from geolocated databases supply background appearance via temporally routed cross-attention; 
and instance tokens registered in a write-once appearance bank preserve individual object identities across chunk boundaries and occlusions through instance-masked cross-attention. 
Although this explicit memory substrate substantially mitigates structural and identity drift, injecting such dense, multi-scale conditioning signals inevitably shifts the generator off its pretrained distribution, 
compromising native visual sharpness and multi-agent physical plausibility. 
To overcome this trade-off, 
we introduce Drive DiffusionNFT, an online post-training framework that optimizes self-generated rollouts via flow-matching reinforcement learning guided by hierarchy-aligned visual rewards and a physical judge. 
Beyond their individual roles, we empirically observe a practical synergy between explicit memory persistence and online policy optimization: 
the invariant memory state appears to confine the effective policy optimization space, coinciding with reduced semantic drift and reward hacking during recursive rollouts, 
while online RL systematically restores fine-grained visual fidelity and enforces physical commonsense without compromising global stability.

Extensive evaluations on the standard nuScenes benchmark and a self-collected challenging long-horizon dataset demonstrate the effectiveness of WAM-Sim. 
On nuScenes, WAM-Sim achieves superior generation fidelity ($\text{FID} = 5.75$, $\text{FVD} = 32.40$) alongside strong downstream perception performance ($22.30$ mAP, $70.32$ road mIoU). 
On minute-scale rollouts of self-collected data, it effectively mitigates quality–horizon decay, preserving high scene similarity ($0.910$) and instance identity coherence ($0.681$) across extended temporal gaps. 
Furthermore, human preference studies and physical evaluations confirm that tightly coupling persistent hierarchical memory with online flow-matching post-training substantially mitigates structural and identity drift, synthesizing realistic and physically sound driving simulations.

\begin{figure}[!t]
  \centering
  \includegraphics[width=1.0\linewidth]{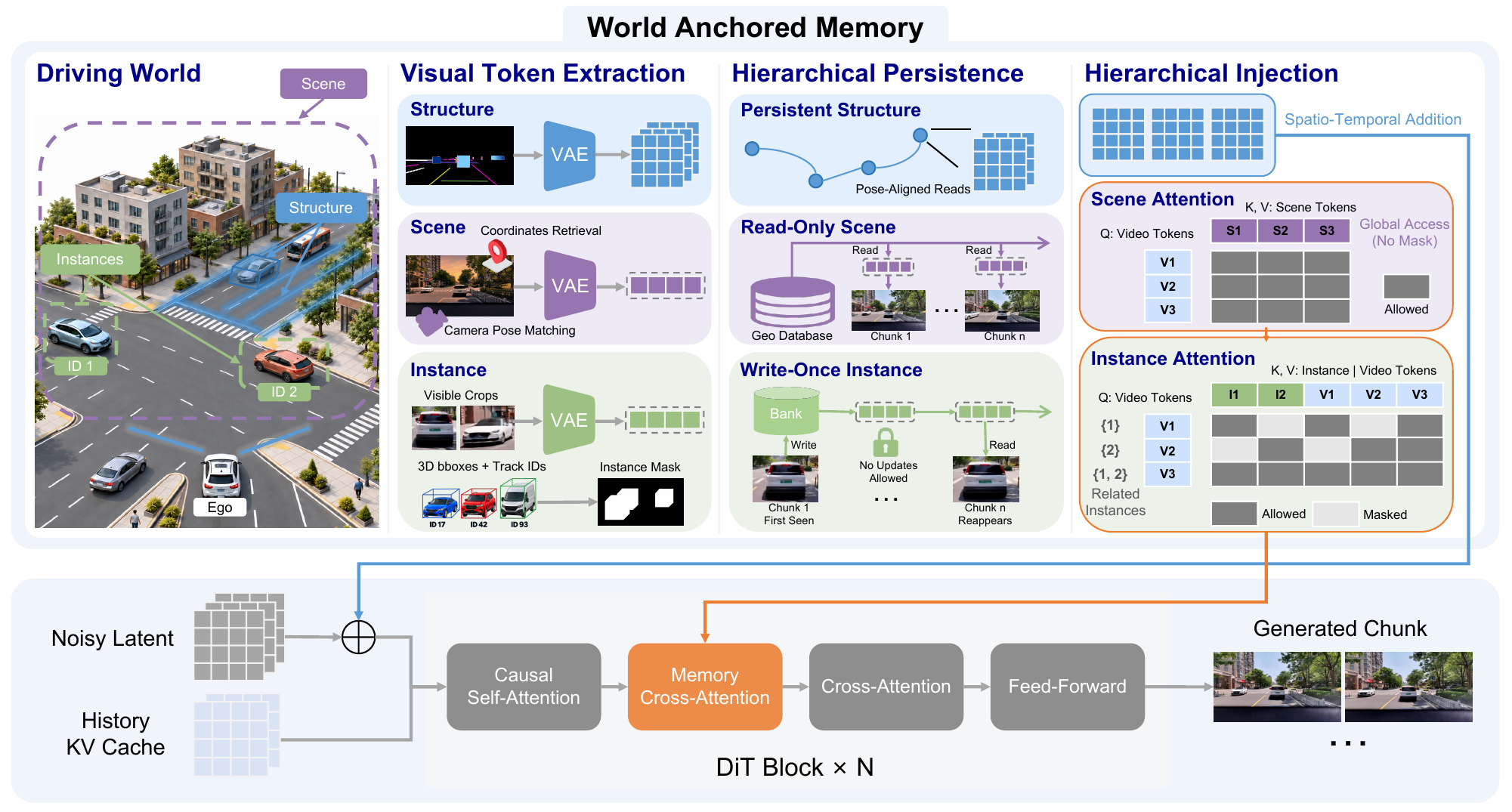}
    \caption{Overview of the WAM-Sim architecture and world anchored memory.
    Persistent context is structured across three temporal and semantic scales: 
    1) Structure: pose-aligned HD-map tokens injected via spatio-temporal addition to anchor road geometry; 
    2) Scene: geolocated background tokens retrieved from a database and accessed via temporally routed scene cross-attention; and 
    3) Instance: object appearance crops stored in a write-once appearance bank and retrieved via instance-masked cross-attention to isolate participant identities. 
    The hierarchical memory state conditions a 3D Diffusion Transformer (DiT) operating on noisy latents and history KV caches to synthesize autoregressive, consistent driving video chunks.}
  \label{fig:framework}
  \vspace{-4mm}
\end{figure}

\section{Related Work}
\label{sec:related}

\textbf{Driving World Models.}
Driving video generation has evolved from action- and trajectory-conditioned models~\citep{hu2023gaia,russell2025gaia,gao2024vista} to layout- and map-conditioned video synthesis~\citep{wang2024drivedreamer,wen2024panacea,gao2024magicdrive,gao2024magicdrivedit,zhao2024drivedreamer2,wang2024drivewm,li2023drivingdiffusion,mei2024dreamforge}, autoregressive long-horizon generation~\citep{jia2023adriver,hu2024drivingworld,zhang2025epona,guo2024infinitydrive}, geometry-aware 4D modeling~\citep{yang2025xscene,dive2025,dist4d2025,unimlvg2025}, and driving-specialized foundation models~\citep{nvidia2025cosmos,ren2025cosmosdrive,ni2025opendwm,basant2026omnidreams}.
InstaDrive~\citep{yang2026instadrive} propagates instance features to preserve identity, yet treats such evidence as transient per-chunk conditioning;
HorizonDrive~\citep{zhang2026horizondrive} extends rollouts toward minute-scale horizons via streaming context.
Across these methods, maps, boxes, and retrieved context remain transient, with no persistent, addressable state binding geographic location and instance identities; 
we replace transient conditioning with a world anchored memory.

\textbf{Memory and Long-Horizon Consistency.}
Persistence in interactive world models is achieved by attention caches~\citep{valevski2024gamengen,decart2024oasis,parkerholder2024genie2,deepmind2025genie3}, retrieval over past observations~\citep{xiao2025worldmem,yu2025contextasmemory,li2025vmem,zhao2026spatia,wu2025longtermspatialmemory}, or geographic grounding~\citep{deng2024streetscapes,wang2025terasimworld,jia2026spatialretrieval}.
Orthogonal efforts extend or compress context~\citep{qiu2024freenoise,kim2024fifo,henschel2024streamingt2v,zhang2025framepack,huang2025memoryforcing,furuta2025worldpack} and mitigate exposure bias~\citep{chen2024diffusionforcing,yin2025causvid,huang2025selfforcing,liu2026rollingforcing}.
Closest to ours, ConsisDrive~\citep{yang2026consisdrive} encodes instance identity and trajectory through instance-masked attention and losses; 
we adopt its mask to specify \emph{where} identity applies, while our write-once appearance bank determines \emph{what} persists across occlusions and chunk boundaries.
Spatial Retrieval~\citep{jia2026spatialretrieval} grounds generation geographically, but retains no persistent scene or instance state.
Each mechanism addresses persistence in isolation; 
we unify geometric anchoring, scene appearance, and instance identity within a single memory substrate.

\textbf{Reinforcement Learning for Visual Generation.}
Reward alignment of visual diffusion models has advanced from policy gradients and preference optimization~\citep{black2023ddpo,fan2023dpok,wallace2023diffusiondpo,xu2023imagereward} to online group-relative flow-matching post-training~\citep{liu2025flowgrpo,xue2025dancegrpo,zheng2025diffusionnft}.
Recently, RL post-training has been applied to video world models: RLVR-World~\citep{wu2025rlvrworld} leverages verifiable rewards for video generation, while Reward Forcing~\citep{lu2025rewardforcing} distills reward signals into streaming autoregressive diffusion rollouts.
Orthogonal to these approaches, which optimize generic video rewards, WAM-Sim is, to the best of our knowledge, the first to combine forward-process DiffusionNFT optimization with memory-aligned hierarchical driving rewards---unifying multi-scale visual fidelity with VLM-guided physical plausibility---to constrain policy search and mitigate structural and quality decay on self-generated multi-chunk rollouts.
\begin{figure}[!t]
  \centering
  \includegraphics[width=1.0\linewidth]{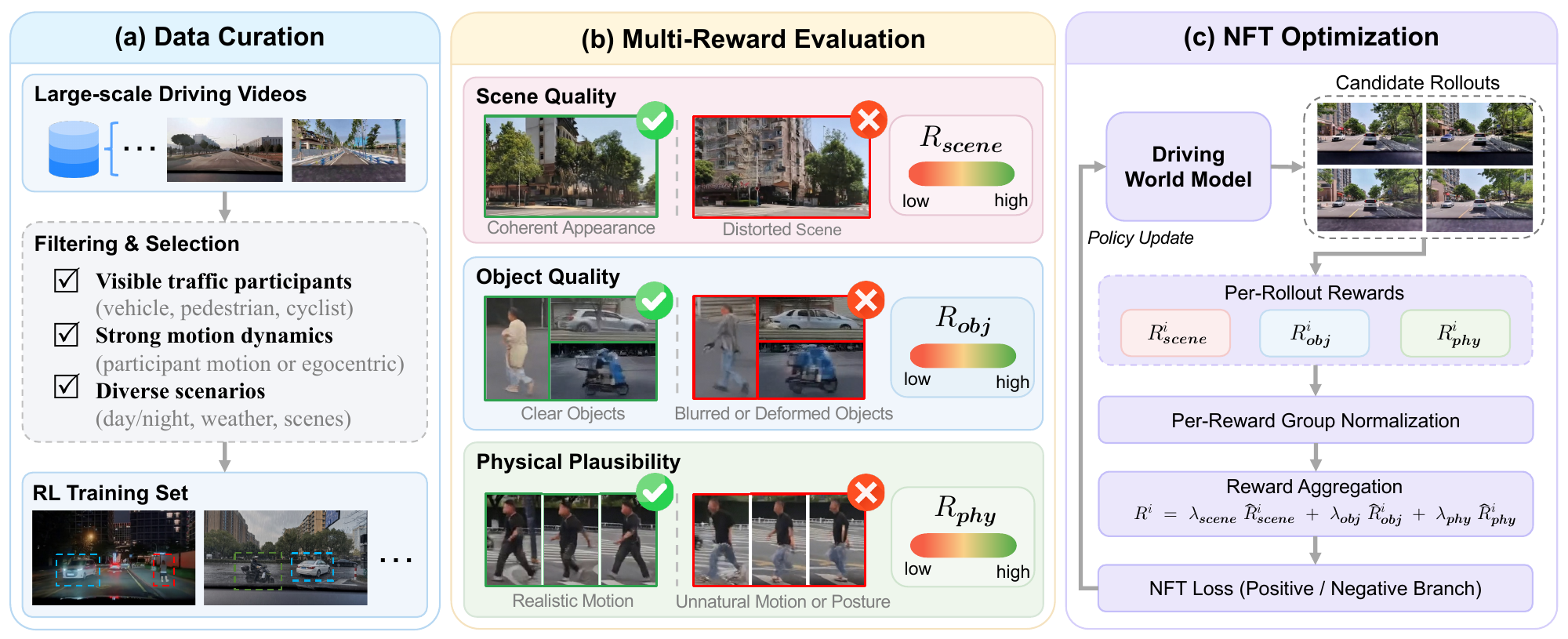}
    \caption{Overview of the Drive DiffusionNFT pipeline.
(a) Data curation constructs a diverse RL training set with visible traffic participants and strong motion dynamics.
(b) Multi-reward evaluation assesses scene quality, object quality, and physical plausibility.
(c) NFT optimization normalizes and aggregates the rewards to update the driving world model.}
  \label{fig:rl_pipeline}
  \vspace{-5mm}
\end{figure}
\vspace{-2mm}
\section{Method}
\label{sec:method}
\vspace{-2mm}
As shown in Figure~\ref{fig:framework}, 
we present a long-horizon driving world model (Section~\ref{sec:method:formulation}) built on two mutually reinforcing components: 
a persistent world anchored memory
that anchors generation to a real geographic location and to per-instance identities (Section~\ref{sec:method:memory}), 
and Drive DiffusionNFT, 
an online post-training stage that optimizes visual and physical realism on self-generated rollouts (Section~\ref{sec:method:rl}). 
The two components are synergistic by design: 
the memory state delimits the manifold within which reward optimization operates, 
while the rewards are aligned with the memory hierarchy, 
so that the policy improves precisely the properties that memory persists.

\subsection{Problem Formulation}
\label{sec:method:formulation}

Given a local HD-map $e_c$, 3D bounding-box agent controls $a_c$, a global ego pose trajectory $P_c$, and a persistent memory state $\mathcal{M}_c$, the model autoregressively synthesizes driving video chunks $\hat{I}_c \in \mathbb{R}^{V \times T \times H \times W \times 3}$ of $T$ frames across $V$ camera views:
\begin{equation}
\hat{I}_{c}
\sim
p_\theta\!\left(
\cdot \,\middle|\,
\hat{I}^{\mathrm{tail}}_{c-1},\,
e_{c},\,a_{c},\,P_{c},\,\mathcal{M}_{c}
\right),
\qquad
\mathcal{M}_{c+1}=\mathrm{Update}(\mathcal{M}_{c},\, e_c, a_c, P_c, \hat{I}_c).
\label{eq:chunked_generation}
\end{equation}
Here, $a_c$ encodes 3D bounding-box layout sequences for surrounding agents, and $P_c$ provides the global ego trajectory used for geolocated reference retrieval. $V$ denotes the number of camera views.
Specifically, $\mathrm{Update}(\cdot)$ is an append-only operation: it registers newly emerging traffic instances into the appearance bank upon initial observation and advances temporal segment indices, 
guaranteeing that existing memory entries are never overwritten or corrupted by self-generated video content.
Long-horizon generation satisfies three requirements:
1) no \emph{structural drift}---the video remains anchored to the road geometry and background appearance;
2) no \emph{identity drift}---each tracked participant preserves its appearance across chunks, occlusion, and re-emergence;
3) no \emph{quality decay}---visual and physical fidelity do not degrade with horizon.

\subsection{World Anchored Memory}
\vspace{-1mm}
\label{sec:method:memory}
A driving world retains information at multiple scales---road topology, geolocated scene context, and traffic-participant identities---which differ in semantics, spatial scope, and temporal persistence.
We therefore formalize the world state as a hierarchical memory tuple $\mathcal{M} = \left( \mathcal{M}^{\mathrm{struct}}, \mathcal{M}^{\mathrm{scene}}, \mathcal{M}^{\mathrm{inst}} \right)$, where each tier is maintained under a dedicated persistence mechanism and retrieved through an operator aligned with its spatio-temporal validity domain.

\textbf{Memory Token Extraction.}
We extract memory tokens across three distinct semantic and temporal scales. 
First, structural tokens capture static road topology by projecting the per-frame HD-map sequence into the latent video grid via a dedicated map embedding layer. 
Second, scene tokens represent persistent background appearance; 
we partition the rollout of $C_{\mathrm{total}}$ frames into $N_b=\lceil C_{\mathrm{total}}/T_b \rceil$ non-overlapping temporal segments of $T_b$ frames (one chunk per segment by default) and, for each segment, retrieve the nearest reference image from the geolocated database using the segment's global ego pose, subject to a spatial radius $r_{\max}$ and a camera-heading alignment constraint; if no reference satisfies both criteria, the segment reuses the reference of its preceding segment. 
Each reference image is VAE-encoded into $L$ patch tokens $S_n \in \mathbb{R}^{L \times D}$, forming the unified scene memory matrix $S = \operatorname{Concat}_{n=1}^{N_b} S_n \in \mathbb{R}^{N_b L \times D}$.
Third, instance tokens encode object identity. For each tracked instance, 
after filtering observations by the bounding-box area range $[\rho_{\min},\rho_{\max}]$, 
we divide the remaining frames into at most $K_f$ temporal groups and select the largest observation from each.
The selected crops are VAE-encoded and projected into $B_i\in\mathbb{R}^{K\times D}$, 
where $K=K_fK_p$ and $K_p$ is the number of tokens per crop. A validity vector $\nu_i$ masks padded slots.

\textbf{Hierarchical Persistence.}
The three memory tiers operate under persistence mechanisms strictly aligned with their temporal dynamics. 
Structural memory is trajectory-aligned: 
the global road topology remains invariant, while its per-frame HD-map representation dynamically transforms relative to the ego pose. 
Scene memory is geographically grounded and read-only: 
geolocated references in $\mathcal{M}^{\mathrm{scene}}$ remain static throughout generation and cannot be overwritten by self-generated content, 
thereby anchoring environmental textures, architecture, and road surfaces to real-world coordinates. 
Instance memory follows a write-once policy: 
upon an agent's initial appearance, its visual tokens $B_i$ are cached in a track-keyed bank $\mathcal{M}^{\mathrm{inst}} = \{\tau_i \mapsto B_i\}$ indexed by non-parametric track identifiers $\tau_i$. 
Locking $B_i$ against subsequent overwrites prevents autoregressive error accumulation from corrupting object identities, substantially improving re-identification across extended occlusions and chunk boundaries.

\textbf{Hierarchical Memory Injection.}
Rather than exposing all memory tokens to every video token, our hierarchical routing progressively conditions a shared video representation with geometry-, environment-, and identity-level information at their corresponding validity domains.

Structural memory provides trajectory-aligned geometric information. Since the HD-map representation is spatially aligned with the noisy video latent, it is incorporated as an additive condition,
$
X^{(0)} = Z_t + \mathcal{M}^{\mathrm{struct}},
\label{eq:structural_injection}
$
where $Z_t$ denotes the noisy video latent at flow timestep $t$. This establishes the geometry-aware video representation on which the subsequent memory routing operates.
Scene memory operates at the temporal-segment level. For video token $j$, let $s(j)$ denote its corresponding segment. The routing mask is defined as $R^{\mathrm{scene}}_{j,(n-1)L+r}=0$ when $s(j)=n$ and otherwise $-\infty$, where $r\in\{1,\ldots,L\}$, ensuring that each token retrieves scene information only from its associated temporal segment.
Scene-memory retrieval is performed on the current geometry-conditioned video representation, producing an updated representation that incorporates the corresponding environmental context.
Instance memory further specializes this shared representation within object-specific spatio-temporal regions. Given $N$ tracked instances, we rasterize their bounding boxes onto the video-token grid to form an instance-membership matrix $\Pi\in\{0,1\}^{L_v\times N}$, where $L_v$ denote the number of video tokens per chunk and $\Pi_{j,i}=1$ if and only if token $j$ belongs to instance $i$.
Only foreground tokens covered by tracked instances participate in the instance-memory branch, while background tokens bypass it. Each foreground token interacts only with video tokens belonging to the same instance and retrieves appearance information only from the corresponding entry in the instance memory bank. Cross-instance connections and invalid memory tokens are masked out. Importantly, instance-memory retrieval operates on the video representation already conditioned by both structural and scene information.

The resulting formulation constitutes a coarse-to-fine hierarchical conditioning process over a shared video representation: structural memory establishes trajectory-aligned geometry, scene memory introduces segment-specific environmental context, and instance memory preserves object-specific appearance within tracked regions. Although their validity domains differ, the three memories jointly form a single hierarchical memory system rather than independent parallel branches. 
\subsection{Drive DiffusionNFT: Synergistic Online Post-Training}
\label{sec:method:rl}

Although explicit memory persistence mitigates structural and identity drift, incorporating multi-scale conditioning introduces distribution shifts that can degrade visual fidelity and physical consistency. 
To resolve this trade-off, we introduce Drive DiffusionNFT, an online post-training framework optimizing self-generated rollouts via flow-matching reinforcement learning. 

\textbf{Hierarchical Reward Formulation.}
As shown in Figure~\ref{fig:rl_pipeline} (b), our rewards mirror the memory hierarchy and the rollout reward $R^i$ for candidate $i$ integrates three multi-level criteria:
\begin{equation}
R^i = \lambda_{\mathrm{scene}}\tilde{R}_{\mathrm{scene}}^i + \lambda_{\mathrm{obj}}\tilde{R}_{\mathrm{obj}}^i + \lambda_{\mathrm{phy}}\tilde{R}_{\mathrm{phy}}^i,
\label{eq:rollout_reward}
\end{equation}
where $\tilde{R}_{m}^{i}$ denotes the group-normalized score of reward component $m$.
Specifically, scene-level perceptual quality averages the VBench Imaging Quality scores over sampled frames \citep{huang2024vbench},
and $R_{\mathrm{scene}}^{i}=\frac{1}{|\mathcal{S}^{i}|}\sum_{f} q_{\mathrm{scene}}(\hat{\mathbf{V}}_f^{i})$.
Object-level recognizability is evaluated on valid agent crops using binary classifiers trained on driving scenes from WorldLens \citep{liang2026worldlens}, covering two categories: vehicles and pedestrians.
To avoid saturation of sigmoid confidence and retain greater score discrimination, we use normalized pre-sigmoid classifier logits,
$R_{\mathrm{obj}}^{i}=\frac{1}{N_{\mathrm{obj}}^{i}}\sum_k \frac{1}{|\mathcal{C}_k^{i}|}\sum_{o\in\mathcal{C}_k^{i}} \tilde{\ell}_{\mathrm{cls}}(o)$,
where $\mathcal{C}_k^{i}$ denotes the valid crops of object $k$, and $\tilde{\ell}_{\mathrm{cls}}(o)$ is the normalized logit from the corresponding class-specific classifier. Crops below a visible-area threshold are excluded.
To explicitly enhance the physical realism of generated driving videos, we apply SoliReward \citep{lian2026solireward} to assess the general physical plausibility and driving scenario subject deformity,
and thereby capturing physically implausible interactions and dynamics in driving scenes, as well as structural or temporal artifacts of traffic participants. 
For each generated video, the physical reward is given directly by its scalar reward score, \(R_{\mathrm{phy}}^{i}=s_{\mathrm{Soli}}(\hat{\mathbf{V}}^{i})\).

\textbf{Flow-Matching Policy Optimization.}
Following DiffusionNFT \citep{zheng2025diffusionnft}, we perform reward-guided online optimization directly over self-generated rollouts. For each condition $\mathbf{c}$, a group of $G$ candidate rollouts is sampled from the collection policy $\pi^{\mathrm{old}}$ with the corresponding clean latents denoted by $\mathbf{x}_0$. For a noisy state $\mathbf{x}_t = \alpha_t \mathbf{x}_0 + \sigma_t \bm{\epsilon}$ with target velocity $\mathbf{v} = \dot{\alpha}_t \mathbf{x}_0 + \dot{\sigma}_t \bm{\epsilon}$ ($\bm{\epsilon} \sim \mathcal{N}(\mathbf{0}, \mathbf{I})$), the flow-matching objective updates policy velocity field $\mathbf{v}_\theta$ via:
\begin{equation}
\mathcal{L}_{\mathrm{NFT}}(\theta) = \mathbb{E}_{\mathbf{c},\,\mathbf{x}_0 \sim \pi^{\mathrm{old}},\,t} \left[ r \left\| \mathbf{v}_\theta^+ - \mathbf{v} \right\|_2^2 + (1 - r) \left\| \mathbf{v}_\theta^- - \mathbf{v} \right\|_2^2 \right],
\label{eq:nft_objective}
\end{equation}
where $\mathbf{v}_\theta^+$ and $\mathbf{v}_\theta^-$ represent positive and negative implicit velocity predictions, and optimality weight $r = \sigma(R^i)$ converts group-relative rollout rewards into relative gradient updates.
\section{Experiments}
\label{sec:experiments}

\vspace{-1mm}
\subsection{Experimental Setups}
\label{sec:exp:setups}
\vspace{-1mm}
\textbf{Implementation Details.}
Our generator is a 3D DiT initialized from pretrained OmniDreams weights and trained in two stages: World-Anchored Memory pre-training and Drive DiffusionNFT post-training. Stage I sequentially optimizes trajectory-aligned structural conditioning, scene memory, and instance memory for 130k, 15k, and 5k steps, respectively. All use batch size 4 and context parallelism degree 8, with learning rates of $1\times10^{-5}$, $3\times10^{-5}$, and $3\times10^{-4}$. Starting from the full Stage I checkpoint, Stage II applies online flow-matching RL to self-generated rollouts using $G=16$ and LoRA adapters (rank 256, LR $1\times10^{-5}$). We use AdamW throughout and train all experiments on 32 Ascend 910B NPUs.

\textbf{Datasets.}
We train and evaluate WAM-Sim on two dataset regimes. 
For standard benchmarking, we evaluate on nuScenes, 
which provides 850 20-second driving scenes (700 for training, 150 for validation) with synchronized multi-view video, 3D bounding boxes, and HD-map annotations. 
To evaluate long-horizon stability, we train our model on a self-collected dataset comprising 121.8K multi-modal clips ($\sim$1,000 hours) across diverse Chinese urban, rural, and highway environments. 
Each training sample includes 10 FPS front-view video ($704{\times}1280$), 
3D bounding boxes with persistent track IDs, 
rendered HD-map topologies, GPS coordinates linked to a geolocated visual reference database, 
and structured captions. 
The corresponding benchmark evaluation set comprises 350 curated long-horizon clips featuring adverse weather, complex maneuvers, vulnerable road users, and extended occlusion–re-emergence events (representative scenarios shown in Fig.~\ref{fig:traffic-participants}).

\textbf{Evaluation Metrics.}
We evaluate generation across four core dimensions: 
1) visual fidelity via FID, FVD, and VBench Quality Score; 
2) long-horizon consistency via Subject Coherence (instance identity) and Scene Similarity (background stability); 
3) object \& physical realism via detection-gated Subject Fidelity and VLM-guided Physical Plausibility; and 
4) downstream perception on nuScenes via 3D detection (mAP, NDS) and BEV segmentation mIoU (road, vehicle). 

\subsection{Main Benchmark Comparisons}
\label{sec:exp:main_results}

\textbf{nuScenes Benchmark.}
Table~\ref{tab:nuscenes_by_paradigm} evaluates multi-view generation fidelity and downstream perception consistency on nuScenes.
Lacking geolocated reference databases, WAM-Sim operates without scene memory in this setting.
At $704{\times}1280$ synthesis resolution, WAM-Sim achieves a visual fidelity ($\text{FID} = 7.07$, $\text{FVD} = 42.01$), outperforming the representative image and video generation paradigms.
Furthermore, this high-fidelity synthesis translates into robust geometric control, yielding competitive downstream 3D detection ($19.67$ mAP) and state-of-the-art BEV road segmentation ($68.08$ mIoU).
For the memory-enabled evaluation, we use the open-source nuScenes-Geography dataset from Spatial Retrieval~\citep{jia2026spatialretrieval} as a geolocated reference database. Enabling scene memory reduces FID from $7.07$ to $5.75$ and FVD from $42.01$ to $32.40$. It also improves mAP and NDS by $2.63$ and $1.06$ points, respectively, and road and vehicle mIoU by $2.24$ and $0.89$ points. These consistent improvements provide public-benchmark evidence for the benefits of memory in both generation fidelity and downstream perception.

\begin{table}[t]
    \centering
    \caption{
      Comparisons of controllable driving scene generation on the
      nuScenes dataset, grouped by generation paradigm.
      $\mathrm{FID}$ is computed over keyframes corresponding to all
      nuScenes validation.
    }
    \vspace{-3mm}
    \label{tab:nuscenes_by_paradigm}
    {
    \renewcommand{\tablefont}{\scriptsize}
    \setlength{\tabcolsep}{12.5pt}
    \renewcommand{\arraystretch}{1.08}
    
    \begin{tabular}{@{}l|c|cc|cc|cc@{}}
      \toprule

      \multirow{2}{*}{\textbf{Method}} &
      \multirow{2}{*}{
        \textbf{\begin{tabular}[c]{@{}c@{}}
          Synthesis\\
          Resolution
        \end{tabular}}
      } &
      \multicolumn{2}{c|}{\textbf{Fidelity}} &
      \multicolumn{2}{c|}{\textbf{3D Detection}} &
      \multicolumn{2}{c}{\textbf{BEV Seg mIoU}} \\

      \cmidrule(lr){3-4}
      \cmidrule(lr){5-6}
      \cmidrule(lr){7-8}

      & &
      FVD $\downarrow$
      & FID $\downarrow$
      & mAP $\uparrow$
      & NDS $\uparrow$
      & Road $\uparrow$
      & Vehicle $\uparrow$ \\

      \midrule

      Ori nuScenes
      & $900 \times 1600$
      & -
      & -
      & 35.54
      & 41.21
      & 73.66
      & 34.82 \\

      \midrule
      \multicolumn{8}{@{}l}{
        \textit{\textbf{(I) Image Generation Methods}}
      } \\
      \midrule

      BEVControl
      & $224 \times 400$
      & -
      & 24.85
      & -
      & -
      & 60.80
      & 26.80 \\

      PerLDiff
      & $256 \times 384$
      & -
      & 13.36
      & 15.24
      & 24.05
      & 61.26
      & 27.13 \\

      DrivePTS
      & $224 \times 480$
      & -
      & 11.45
      & 15.37
      & 25.49
      & 63.95
      & 27.82 \\

      MagicDrive
      & $224 \times 400$
      & -
      & 16.20
      & 12.30
      & 23.32
      & 61.05
      & 27.01 \\

      \midrule
      \multicolumn{8}{@{}l}{
        \textit{\textbf{(II) Video Generation Methods}}
      } \\
      \midrule

      DriveDreamer
      & $256 \times 448$
      & 452.0
      & 26.80
      & -
      & -
      & -
      & - \\

      DriveDreamer-2
      & $256 \times 448$
      & 105.10
      & 25.00
      & -
      & -
      & -
      & - \\

      DrivingDiffusion
      & $512 \times 512$
      & 332.0
      & 15.83
      & -
      & \textbf{33.10}
      & 63.20
      & 31.60 \\

      Panacea
      & $256 \times 512$
      & 139.0
      & 16.96
      & 11.58
      & 22.31
      & 55.78
      & 22.74 \\

      DreamForge
      & $224 \times 400$
      & 224.8
      & 14.61
      & 13.01
      & 22.16
      & 65.27
      & 28.36 \\

      MagicDrive-V2
      & $848 \times 1600$
      & 94.84
      & 20.91
      & 17.65
      & -
      & 59.79
      & \textbf{32.73} \\

      \midrule

      Ours w/o mem.
      & $704 \times 1280$
      & 42.01
      & 7.07
      & 19.67
      & 29.95
      & 68.08
      & 31.53 \\

      \textbf{Ours w/ mem.}
      & $704 \times 1280$
      & \textbf{32.40}
      & \textbf{5.75}
      & \textbf{22.30}
      & 31.01
      & \textbf{70.32}
      & 32.42 \\

      \bottomrule
    \end{tabular}
    }
    \vspace{-3mm}
\end{table}

\begin{table}[!t]
\centering
\footnotesize
\caption{Evaluation on the Self-Collected Dataset. All methods generate videos conditioned on ego trajectories and agent bounding boxes.}
\vspace{-3mm}
\label{tab:long_horizon_main}
{
\renewcommand{\tablefont}{\scriptsize}
\setlength{\tabcolsep}{4pt}
\renewcommand{\arraystretch}{1.12}
\begin{tabular}{l|cc|ccc|cc|cc}
\toprule
\multirow{2}{*}{\textbf{Method}} & \multicolumn{2}{c|}{\textbf{Visual Fidelity}} & \multicolumn{3}{c|}{\textbf{VBench Metrics}} & \multicolumn{2}{c|}{\textbf{Persistence}} & \multicolumn{2}{c}{\textbf{Object \& Physics}} \\
\cmidrule(lr){2-3} \cmidrule(lr){4-6} \cmidrule(lr){7-8} \cmidrule(lr){9-10}
& FID $\downarrow$ & FVD $\downarrow$ & Quality $\uparrow$ & Subject $\uparrow$ & Background $\uparrow$ & Sub. Coh. $\uparrow$ & Scene Sim. $\uparrow$ & Sub. Fid. $\uparrow$ & Phys. Plaus. $\uparrow$ \\
\midrule
\multicolumn{10}{@{}l}{
\textit{\textbf{(I) Inference Only}}
      } \\
\midrule
UniMLVG & 44.21&753.16&75.75&83.43&89.77&0.591&0.708&39.19&46.57 \\
OpenDWM & 37.69&535.42&76.54&82.36&90.11&0.532&0.731&37.24&47.44 \\
HorizonDrive & 24.86&218.51&78.41&84.55&88.09&0.553&0.804&42.98&51.19 \\
\midrule
\multicolumn{10}{@{}l}{
\textit{\textbf{(II) SFT on Self-Collected Data}}
      } \\
\midrule
OmniDreams & 11.58&81.64&77.63&86.17&92.35& 0.646 &0.863&42.16&51.45 \\
\textbf{Ours} & \textbf{10.95} & \textbf{74.20} & \textbf{78.71} & \textbf{86.98} & \textbf{92.54} & \textbf{0.681} & \textbf{0.910} & \textbf{44.00} & \textbf{55.96} \\
\bottomrule
\end{tabular}%
 }
 \vspace{-4mm}
\end{table}

\begin{table}[!t]
\centering

\begin{minipage}[t]{0.55\linewidth}
    \centering
    \captionof{table}{
    World memory component ablation on the Self-Collected Dataset. Scene and Instance Memory improve generation consistency.
    }
    \vspace{-3.5mm}
    \label{tab:system_ablation}

    \setlength{\tabcolsep}{8pt}
    \renewcommand{\arraystretch}{1.0}

    \begin{tabular}{cc|ccc}
    \toprule
    \textbf{Scene}
    & \textbf{Instance}
    & \textbf{Sub. Coh.} $\uparrow$
    & \textbf{Scene Sim.} $\uparrow$
    & \textbf{FVD} $\downarrow$ \\
    \midrule
    -- & -- & 0.646 & 0.863 & 81.64 \\
    $\surd$ & -- & 0.659 & \textbf{0.911} & 76.68 \\
    -- & $\surd$ & 0.680 & 0.874 & 85.55 \\
    $\surd$ & $\surd$ &\textbf{0.681} & 0.910 & \textbf{74.20} \\
    \bottomrule
    \end{tabular}%
\end{minipage}
\hfill
\begin{minipage}[t]{0.43\linewidth}
    \centering
    \captionof{table}{
    Robustness to scene-memory dropout. Each reference is independently dropped at inference with probability $p$.
    }
    \vspace{-0.8mm}
    \label{tab:scene_memory_robustness}

    \setlength{\tabcolsep}{11pt}
    \renewcommand{\arraystretch}{1.0}

    \begin{tabular}{c|ccc}
    \toprule
    \textbf{Drop Prob. $p$}
    & \textbf{Scene Sim.} $\uparrow$
    & \textbf{FVD} $\downarrow$ \\
    \midrule
    $0.0$ & \textbf{0.910} & \textbf{74.20} \\
    $0.1$ & 0.908 & 77.03 \\
    $0.3$ & 0.904 & 78.60 \\
    \bottomrule
    \end{tabular}%
\end{minipage}
\vspace{-4mm}
\end{table}

\begin{figure}[!t]
    \centering
    \includegraphics[width=\linewidth]{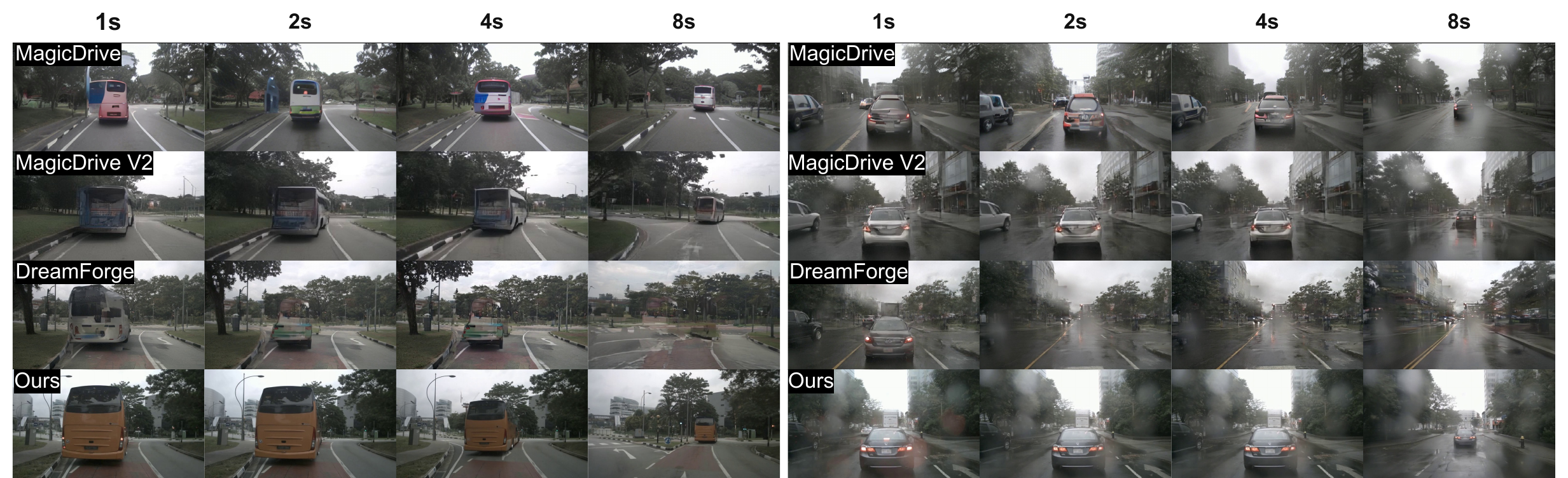}
    \vspace{-6mm}
    \caption{Qualitative comparison results on nuScenes video generation.}
    \label{fig:nuscenes-qualitative}
    \vspace{-5mm}
\end{figure}

\begin{figure}[!t]
  \centering

  \begin{minipage}[t]{0.585\linewidth}
    \centering
    \includegraphics[width=\linewidth]{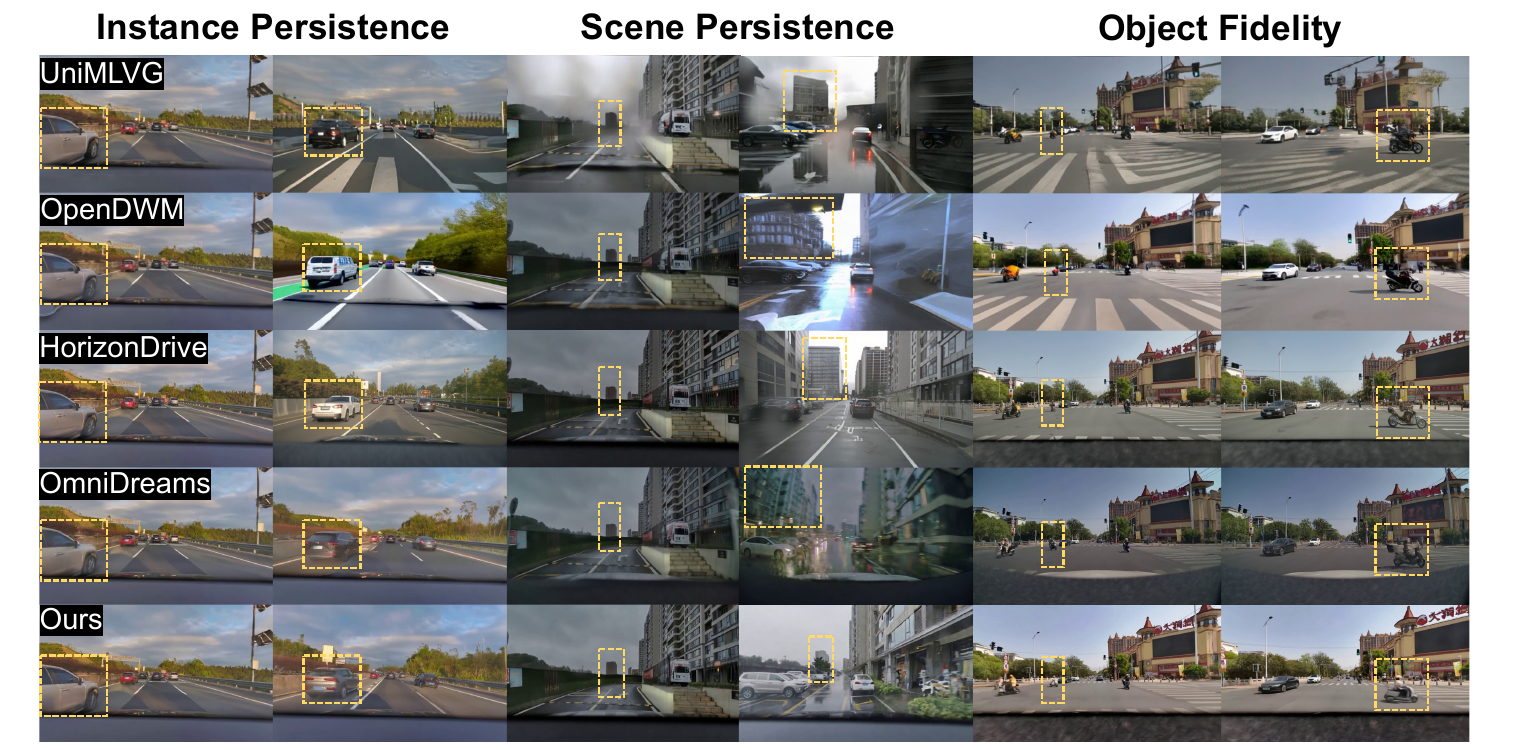}
    \vspace{-7mm}
    \caption{Qualitative comparison results on the self-collected dataset.}
    \label{fig:self-collected}
  \end{minipage}
  \hfill
  \begin{minipage}[t]{0.385\linewidth}
    \centering
    \includegraphics[width=\linewidth]{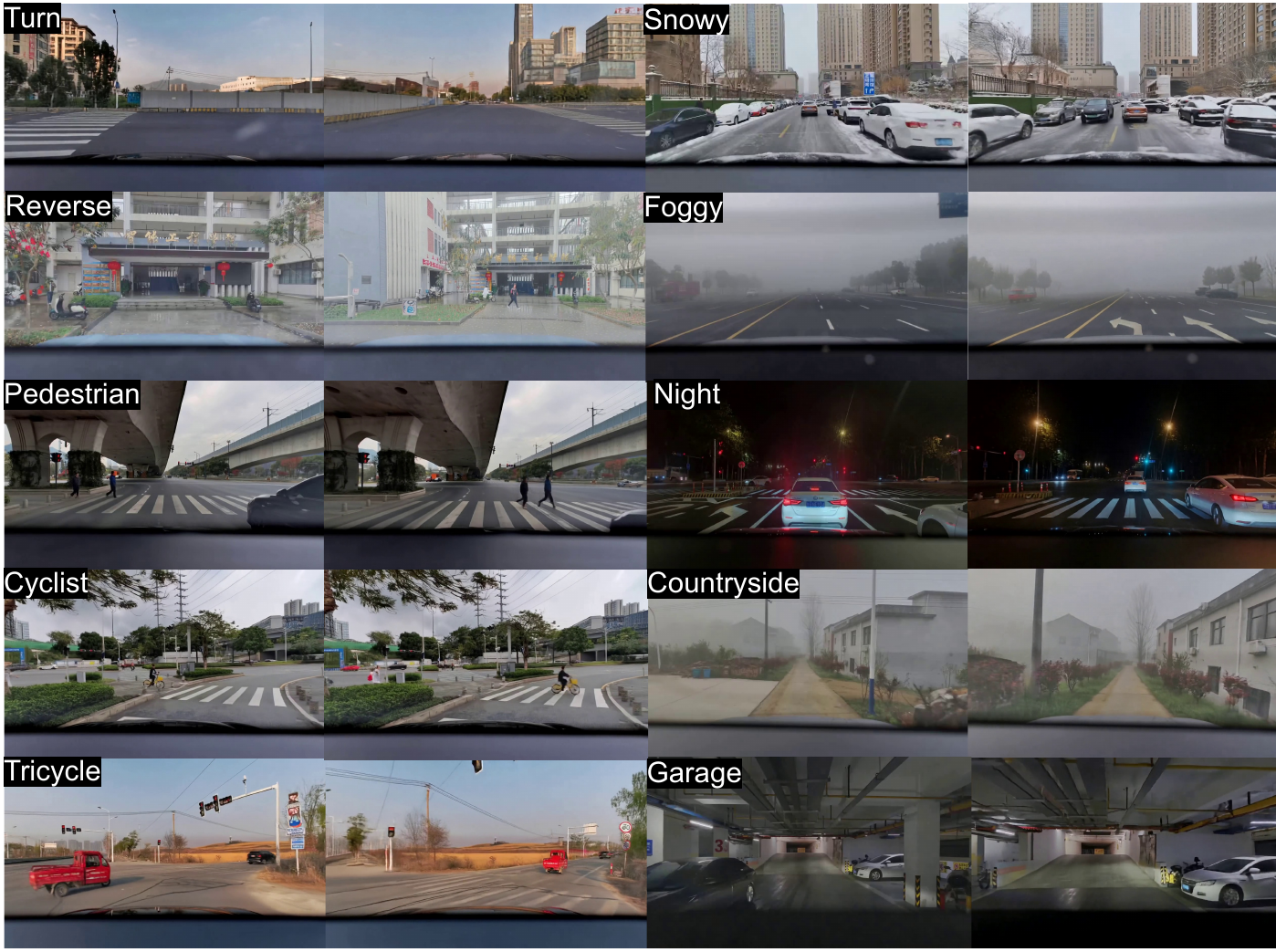}
    \vspace{-7mm}
    \caption{Diverse generation results on our
    self-collected dataset.}
    \label{fig:traffic-participants}
  \end{minipage}

  \vspace{-6mm}
\end{figure}

\textbf{Self-Collected Dataset.} 
As shown in Table~\ref{tab:long_horizon_main} and Figure~\ref{fig:traffic-participants}, by anchoring rollouts with persistent world memory, WAM-Sim effectively mitigates quality–horizon decay, 
outperforming other baselines with $10.95$ FID and $74.20$ FVD while securing superior background stability ($0.910$ Scene Similarity) and identity retention ($0.681$ Subject Coherence). 
Furthermore, Drive DiffusionNFT online post-training successfully enforces physical commonsense and object realism, yielding peak VBench Quality ($78.71$), Subject Fidelity ($44.00$), and Physical Plausibility ($55.96$).

\subsection{Ablation and Discussion}
\label{sec:exp:ablations}

\textbf{Architecture Component Ablation.}
Table~\ref{tab:system_ablation} disentangles the complementary roles of Scene Memory and Instance Memory on the self-collected dataset. 
Individually, Scene Memory anchors global environmental context, 
boosting Scene Similarity from $0.863$ to $0.911$ and reducing FVD from $81.64$ to $76.68$ over the memory-free baseline. 
Conversely, Instance Memory strengthens fine-grained identity persistence, elevating Subject Coherence to $0.680$, though its standalone FVD degrades to $85.55$ due to a lack of global structural bounds. 
Jointly integrating both tiers achieves optimal performance—yielding the lowest FVD, near-peak Subject Coherence, and Scene Similarity. 

\begin{figure}[t]
    \centering

    \begin{subfigure}[t]{0.39\linewidth}
        \centering
        \includegraphics[width=\linewidth]{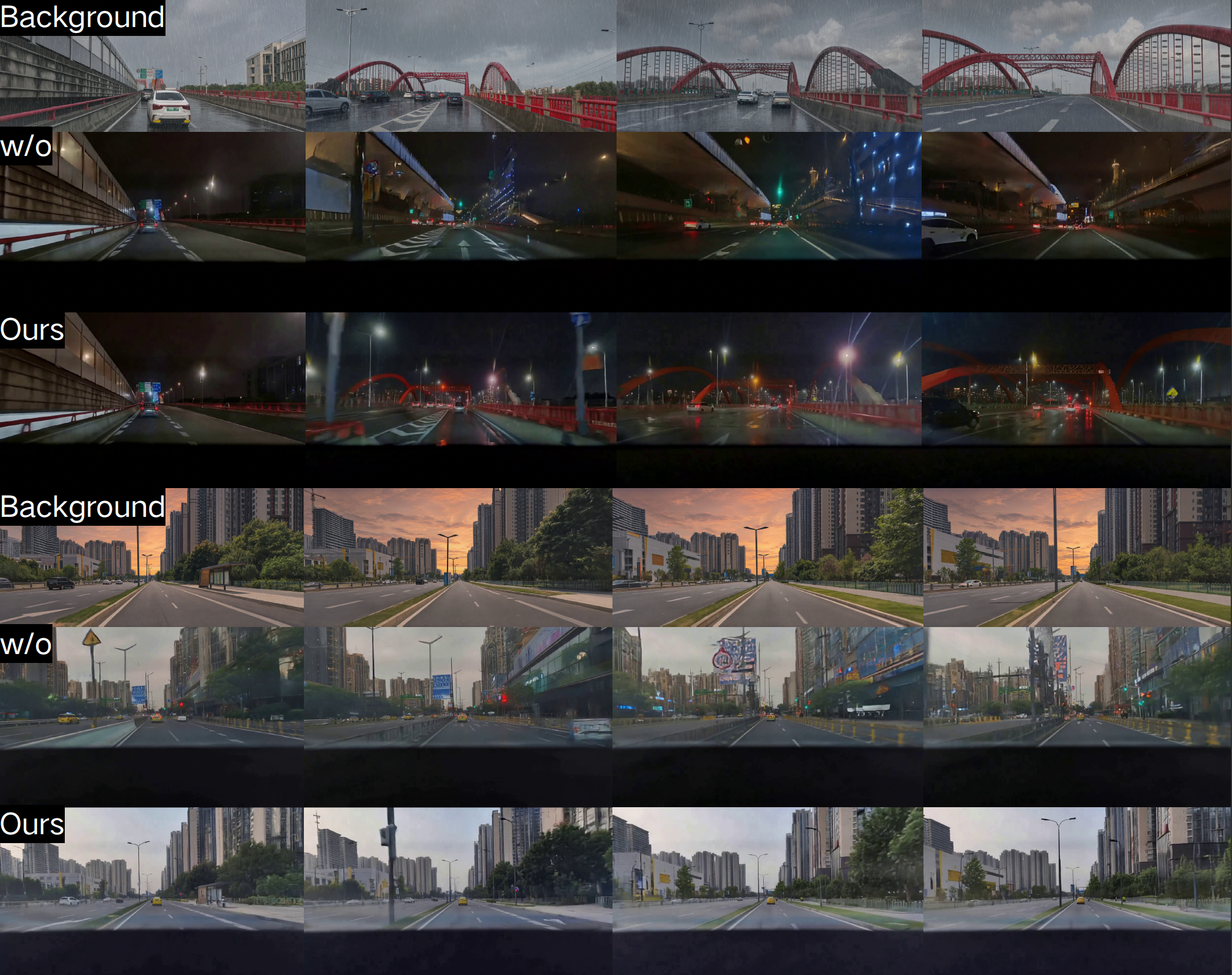}
        \label{fig:bg_id_preservation}
    \end{subfigure}
    \hfill
    \begin{subfigure}[t]{0.60\linewidth}
        \centering
        \includegraphics[width=\linewidth]{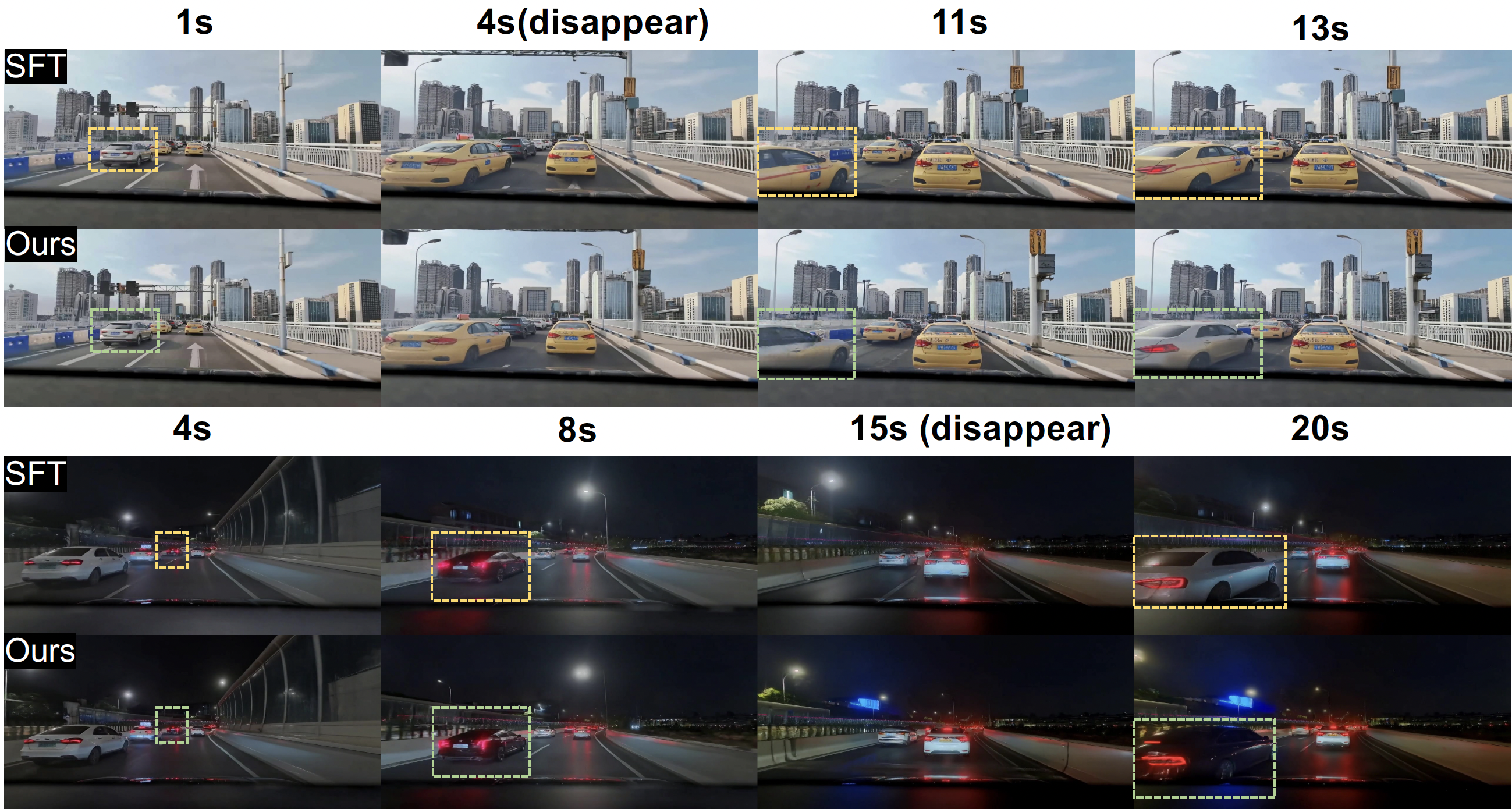}
        \label{fig:id_preservation}
    \end{subfigure}
    \vspace{-6mm}
    \caption{Qualitative comparison of scene- and instance-level persistence. WAM-Sim better preserves background appearance and object identity over long rollouts.}
    \label{fig:memory_qualitative}
    \vspace{-4mm}
\end{figure}

\textbf{Scene Memory.} 
We evaluate the robustness and appearance-layout disentanglement of scene memory.
Quantitatively, introducing reference dropout at inference demonstrates both active reliance and strong fault tolerance (Table~\ref{tab:scene_memory_robustness}); 
even at $p=0.3$, Scene Similarity decays marginally from $0.910$ to $0.904$. 
Qualitatively, Figure~\ref{fig:memory_qualitative}(left) demonstrates that scene memory effectively disentangles persistent spatial layouts from transient environmental conditions: 
when conditioned on a dusk reference image (``Background''), WAM-Sim faithfully preserves the underlying architectural geometry and road layout throughout target daytime rollouts (``Ours''). 
Conversely, omitting scene memory (``w/o'') leads to severe structural drift and hallucinated background geometry.

\textbf{Instance Memory.}
We evaluate the capacity of write-once instance memory to maintain participant identity across extended horizons and occlusion–re-emergence events (Appendix~\ref{app:identity_metrics}). 
We benchmark against the SFT baseline and ConsisDrive, which incorporates instance-level masking without a persistent memory state. 
In Table~\ref{tab:instance_identity}, 
while ConsisDrive increases average identity preservation from 0.642 to 0.680 over SFT via spatial masking alone, 
WAM-Sim surpasses both methods across all metrics. 
Because WAM-Sim and ConsisDrive share the same masking backbone, 
this performance gain isolates the effect of our write-once appearance bank, 
demonstrating that persistent memory prevents autoregressive identity drift across chunk boundaries. 
Qualitatively (Fig.~\ref{fig:memory_qualitative}, right), 
when an agent re-emerges post-occlusion, SFT suffers appearance mutation, 
whereas WAM-Sim recovers its original visual identity by retrieving stored track-keyed features.

\begin{figure}[t]
  \centering

  \includegraphics[width=\linewidth]{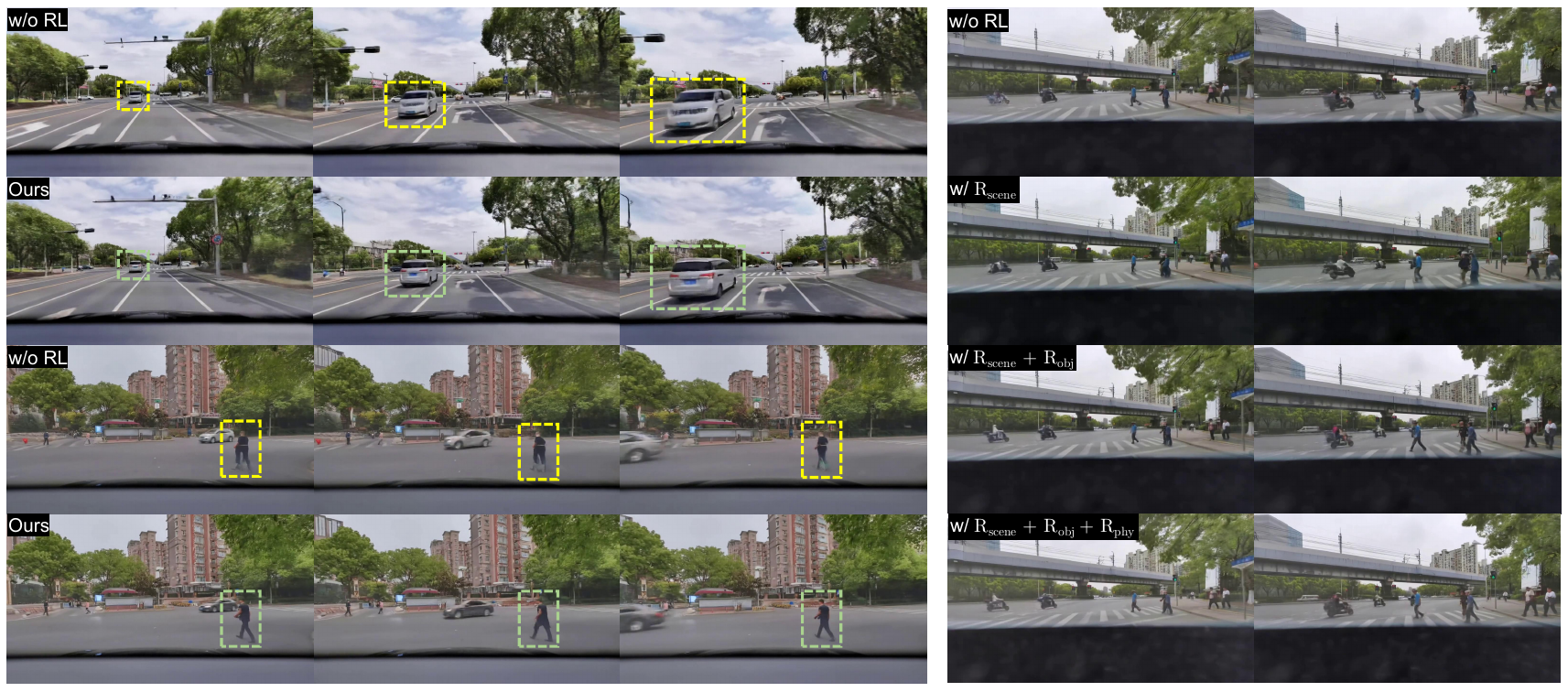}
  \vspace{-6mm}

  \caption{Qualitative ablation of online RL.
Left: comparison between the world-memory model without RL and the final model.
Right: cumulative reward ablation on the same scenario, progressively adding $R_{\mathrm{scene}}$, $R_{\mathrm{obj}}$, and $R_{\mathrm{phy}}$.}
  \label{fig:ablation_rl_reward}
  \vspace{-3mm}
\end{figure}

\begin{table}[!t]
\centering
\caption{Comparison of instance identity preservation on the test set.}
\label{tab:instance_identity}
\vspace{-1mm}
\small
\setlength{\tabcolsep}{7pt}
\vspace{-2mm}
\begin{tabular}{l|cccccccc}
\toprule
\multirow{2}{*}{\textbf{Method}}
& \multicolumn{3}{c}{Long-range Coherence}
& \multicolumn{4}{c}{Re-emergence}
& \multirow{2}{*}{Avg.$\uparrow$} \\
\cmidrule(lr){2-4}
\cmidrule(lr){5-8}

& SC@1S$\uparrow$
& SC@2S$\uparrow$
& SC@3S$\uparrow$
& Sim.$\uparrow$
& Pass Rate$\uparrow$
& Occ-ID@1S$\uparrow$
& Occ-ID@2S$\uparrow$
& \\

\midrule

Baseline (SFT)
& 0.605
& 0.579
& 0.546
& 0.780
& 0.461
& 0.781
& 0.743
& 0.642 \\

ConsisDrive
& \underline{0.634}
& \underline{0.604}
& \underline{0.597}
& \underline{0.805}
& \underline{0.510}
& \underline{0.799}
& \underline{0.812}
& \underline{0.680} \\

\textbf{Ours}
& \textbf{0.653}
& \textbf{0.620}
& \textbf{0.613}
& \textbf{0.810}
& \textbf{0.535}
& \textbf{0.804}
& \textbf{0.833}
& \textbf{0.695} \\

\bottomrule
\end{tabular}
\vspace{-3mm}
\end{table}

\begin{table}[!t]
\centering
\caption{Ablation of reward components in Drive DiffusionNFT post-training.}
\label{tab:rl_reward_ablation}
\setlength{\tabcolsep}{8pt}
\renewcommand{\arraystretch}{1.12}
\vspace{-3mm}
\begin{tabular}{ccc|ccc|cc}
\toprule
$R_{\text{scene}}$ & 
$R_{\text{obj}}$ & 
$R_{\text{phy}}$ & 
\textbf{Img. ($R_{\mathrm{scene}}$)} $\uparrow$ & 
\textbf{Classi. ($R_{\mathrm{obj}}$)} $\uparrow$ & 
\textbf{Phys. ($R_{\mathrm{phy}}$)} $\uparrow$ & 
\textbf{Sub. Fid.} $\uparrow$ & 
\textbf{Phys. Plaus.} $\uparrow$ \\
\midrule
-- & -- & -- & 59.55 & -0.012 & 3.770 & 41.74 & 50.91 \\
$\surd$ & -- & -- & 62.68 & 0.064 & 3.727 & 42.25 & 51.13 \\
$\surd$ & $\surd$ & -- & 61.59 & 0.273 & 3.789 & 43.97 & 53.54 \\
$\surd$ & $\surd$ & $\surd$ & 61.48 & 0.271 & 4.052 & 44.00 & 55.96 \\
\bottomrule
\end{tabular}%
\vspace{-3mm}
\end{table}

\textbf{Drive DiffusionNFT.} 
Table~\ref{tab:rl_reward_ablation} validates the hierarchical reward design in Drive DiffusionNFT. 
Incrementally incorporating $R_{\mathrm{scene}}$, $R_{\mathrm{obj}}$, and $R_{\mathrm{phy}}$ systematically boosts all optimization signals—improving VBench Imaging Quality ($59.55 \to 62.68$), Object Classification score ($-0.012 \to 0.271$), 
and SoliReward Physics score ($3.770 \to 4.052$). 
Crucially, these gains directly transfer to held-out evaluation metrics, 
elevating Subject Fidelity ($41.74 \to 44.00$) and Physical Plausibility ($50.91 \to 55.96$). 
This shows that multi-level reward alignment enhances physical realism and instance fidelity without collapsing into reward hacking. 
Figure~\ref{fig:ablation_rl_reward} (right) qualitatively demonstrates the cumulative visual and physical gains across each reward tier.

\textbf{Synergy Between Memory Bounds and Online RL.} 
As shown in Figure~\ref{fig:ablation_rl_reward} (left), 
explicit memory persistence and online policy optimization are mutually reinforcing: 
world anchored memory bounds the policy optimization manifold to prevent semantic drift across long rollouts, 
while Drive DiffusionNFT refines fine-grained visual sharpness and enforces physically plausible agent motions within this persistent world scaffold.

\textbf{Memory and RL Interaction.}
We report the complete memory on/off × RL on/off comparison, with all variants trained under the budget. Policy-space drift is measured as the DINOv2 Distance between rollouts generated before and after RL. During RL, we update only the self-attention LoRA parameters of the DiT and keep the memory cross-attention and all other modules frozen. In the factorial comparison, we empirically observe that memory helps limit policy drift to some extent: under identical RL settings, the model moves noticeably less from its pre-RL behavior when memory is present (DINOv2 Distance of 0.087 vs. 0.134). At the same time, with memory, RL achieves larger gains on held-out metrics (+2.26 vs. +1.33 on Subject Fidelity, +5.05 vs. +3.21 on Physical Plausibility), indicating that the optimization leads to real quality gains instead of reward hacking. RL also preserves the persistence gains from memory, even improving them slightly. Overall, these results reflect an interaction benefit in which each component reinforces rather than offsets the other.
\begin{table*}[t]
\centering
\caption{Controlled comparison of memory and online RL on the self-collected dataset. All variants share the same backbone, data, and sampling protocol.}
\vspace{-3mm}
\label{tab:memory_rl_factorial}
\small
\setlength{\tabcolsep}{8pt}
\renewcommand{\arraystretch}{1.08}
\resizebox{\textwidth}{!}{%
\begin{tabular}{lccccc}
\toprule
& \multicolumn{2}{c}{Persistence}
& \multicolumn{2}{c}{Object \& Physics}
& \multicolumn{1}{c}{Policy-space Drift} \\
\cmidrule(lr){2-3}\cmidrule(lr){4-5}\cmidrule(lr){6-6}
\textbf{Method}
& \textbf{Sub. Coh.} $\uparrow$
& \textbf{Scene Sim.} $\uparrow$
& \textbf{Sub. Fid.} $\uparrow$
& \textbf{Phys. Plaus.} $\uparrow$
& \makecell{\textbf{DINOv2 Dist.} $\downarrow$\\\textbf{to Pre-RL}} \\
\midrule
Base        & 0.646 & 0.863 & 42.16 & 51.45 & -- \\
RL only     & 0.652 & 0.861 & 43.49 & 54.66 & 0.134 \\
Memory only & 0.678 & 0.905 & 41.74 & 50.91 & -- \\
Memory + RL & \textbf{0.681} & \textbf{0.910} & \textbf{44.00} & \textbf{55.96} & \textbf{0.087} \\
\bottomrule
\end{tabular}%
}
\vspace{-3mm}
\end{table*}

\textbf{Efficiency Analysis.}
Table~\ref{tab:efficiency_tradeoff} compares WAM-Sim with the trajectory-conditioned baseline at the same resolution. WAM-Sim incurs a $13.3\%$ reduction in throughput and a $3.8\%$ increase in peak memory usage, while achieving $51.9\%$ lower FVD over minute-scale rollouts and substantially improving Scene Similarity and Subject Coherence. This trade-off demonstrates the value of persistent memory for long-horizon simulation.

\begin{table*}[t]
\centering
\caption{Efficiency and long-horizon quality at $1280{\times}704$ resolution. WAM-Sim substantially improves minute-scale consistency with modest throughput and memory overhead.}
\vspace{-3mm}
\label{tab:efficiency_tradeoff}
\small
\setlength{\tabcolsep}{7pt}
\renewcommand{\arraystretch}{1.08}
\resizebox{\textwidth}{!}{%
\begin{tabular}{lccccccc}
\toprule
\textbf{Method}
& \textbf{Resolution}
& \textbf{Params}
& \textbf{Throughput} $\uparrow$
& \textbf{VRAM} $\downarrow$
& \textbf{FVD} $\downarrow$
& \textbf{Scene Sim.} $\uparrow$
& \textbf{Sub. Coh.} $\uparrow$ \\
\midrule
Trajectory-conditioned
& $1280{\times}704$
& 2B
& \textbf{8.00 FPS}
& \textbf{45.71}
& 286.54
& 0.697
& 0.322 \\
\textbf{Ours}
& $1280{\times}704$
& 3.1B
& 6.94 FPS
& 47.45
& \textbf{137.92}
& \textbf{0.828}
& \textbf{0.547} \\
\bottomrule
\end{tabular}%
}
\vspace{-5mm}
\end{table*}

\section{Conclusion}
We presented WAM-Sim, a long-horizon driving world model that systematically mitigates structural drift, identity mutation, and quality decay. 
By unifying a persistent world anchored memory—anchoring road topology, geolocated appearance, and instance identity—with Drive DiffusionNFT online post-training, 
WAM-Sim establishes a persistent world state while enforcing visual sharpness and physical plausibility. 
Benchmark evaluations on nuScenes and self-collected driving datasets demonstrate superior generation fidelity, robust downstream perception, and temporal stability over extended rollouts. 
This synergy between invariant memory manifolds and reward-guided policy optimization offers a promising paradigm toward closed-loop simulation.
\subsection*{AI Use Statement}

Generative AI tools were used to assist with language editing, LaTeX formatting, and code debugging. They were not used to generate experimental data or scientific conclusions. All claims, citations, implementations, and reported results were reviewed and verified by the authors, who take full responsibility for the content of this work.

\subsection*{Ethics Statement}

We use nuScenes in accordance with its license and employ a self-collected driving dataset solely for research purposes. Any public release will include only data permitted for redistribution and will exclude or anonymize faces, license plates, sensitive location information, and other personally identifiable content. The human evaluation collects only perceptual judgments of generated videos and reports results in aggregate. Generated videos and automated evaluation scores are intended for research and should not be used for safety-critical decision-making.

\subsection*{Reproducibility Statement}

To ensure full reproducibility, we release our complete training and evaluation code, model checkpoints, and dataset processing pipelines. Detailed architectural configurations, optimization hyperparameters, and hardware environments are thoroughly documented in Appendices~\ref{app:arch} and~\ref{app:data}.

\bibliography{iclr2027_conference}
\bibliographystyle{iclr2027_conference}

\clearpage
\appendix
\begingroup
\renewcommand{\tablefont}{\footnotesize}
\setlength{\emergencystretch}{1em}
\renewcommand{\topfraction}{0.85}
\renewcommand{\bottomfraction}{0.75}
\renewcommand{\textfraction}{0.10}
\renewcommand{\floatpagefraction}{0.75}
\captionsetup{font=small,skip=6pt}
\setlength{\textfloatsep}{14pt plus 3pt minus 2pt}
\setlength{\floatsep}{10pt plus 2pt minus 2pt}
\setlength{\intextsep}{12pt plus 2pt minus 2pt}
\FloatBarrier
\section{Extended Related Work}
\label{app:related}

\textbf{Driving World Models.}
Controllable driving video generation was pioneered by action-conditioned
world models such as GAIA-1 and GAIA-2~\citep{hu2023gaia,russell2025gaia},
high-fidelity trajectory-conditioned prediction in Vista~\citep{gao2024vista},
and multi-view generators conditioned on BEV layouts, HD maps, and 3D
bounding boxes~\citep{wang2024drivedreamer,wen2024panacea,gao2024magicdrive,gao2024magicdrivedit,zhao2024drivedreamer2}.
Autoregressive token-based approaches extend such models to longer
horizons~\citep{jia2023adriver,hu2024drivingworld,zhang2025epona}, while
geometry-aware methods improve spatial and 4D
consistency~\citep{yang2025xscene,dive2025,dist4d2025,unimlvg2025}.
Two recent trends are most relevant to our setting. First, world
foundation models specialized for driving---Cosmos and
Cosmos-Drive-Dreams~\citep{nvidia2025cosmos,ren2025cosmosdrive}, and most
recently OmniDreams~\citep{basant2026omnidreams}---achieve photorealistic,
interactive, real-time rollouts under world-scenario control, with
OmniDreams attaining minute-long closed-loop simulation through a
streaming KV cache and self-forcing-style distillation. Second, a line of
work explicitly targets long driving sequences: InfinityDrive introduces
memory injection and retention for minute-scale driving
video~\citep{guo2024infinitydrive}. Despite this progress, maps, boxes,
and retrieved context are re-specified per chunk and therefore constrain
only instantaneous generation; what persists across chunks---an
addressable state binding a real location and per-instance
identities---remains implicit (e.g., in attention caches) rather than
explicit and controllable. We upgrade transient conditioning to a
persistent world anchored memory while retaining minute-scale
intervenability.

\textbf{Memory and Long-Horizon Consistency.}
Interactive world models increasingly rely on explicit memory. Genie~3
exhibits emergent scene memory over minute-scale rollouts but, by its own
account, cannot reproduce real-world locations with geographic
accuracy~\citep{deepmind2025genie3}. Retrieval-based memories index past
observations for consistency: WorldMem maintains a retrievable bank
indexed by camera pose~\citep{xiao2025worldmem}, VMem indexes past views
by surfel geometry~\citep{li2025vmem}, Spatia preserves an updatable 3D
point cloud maintained with visual SLAM~\citep{zhao2026spatia}, and video
world models with long-term spatial memory support view
re-identification over revisits~\citep{wu2025longtermspatialmemory}.
Complementary to these generic mechanisms, geolocated generation grounds
synthesis in real places: Streetscapes synthesizes street views of an
on-the-fly city rather than actual locations~\citep{deng2024streetscapes};
TeraSim-World grounds safety-critical data synthesis at arbitrary
coordinates but channels real-world appearance through a VLM text
bottleneck restricted to static elements~\citep{wang2025terasimworld};
and, concurrently with our work, Spatial-RAG-AD retrieves geographic
imagery along the ego route and injects it as persistent cross-attention
context into driving video DiTs~\citep{jia2026spatialretrieval}. On the
training side, long-horizon methods either extend context windows
(FreeNoise~\citep{qiu2024freenoise}, FIFO-Diffusion~\citep{kim2024fifo},
StreamingT2V~\citep{henschel2024streamingt2v}) or close the exposure-bias
gap with self-generated rollouts, from Diffusion
Forcing~\citep{chen2024diffusionforcing} and CausVid~\citep{yin2025causvid}
to Self-Forcing~\citep{huang2025selfforcing} and Rolling
Forcing~\citep{liu2026rollingforcing}; OmniDreams concurrently adopts
self-forcing distillation for real-time driving
rollouts~\citep{basant2026omnidreams}. Instance-level consistency within
single clips is pursued by identity-preserving driving
generation~\citep{yang2026consisdrive} and reference-image subject
consistency methods~\citep{shen2026identity}. Existing works, however,
provide either scene-level memory without real geographic anchors,
geographic grounding without a persistent cross-chunk memory state, or
instance consistency without occlusion-surviving, write-once identity
evidence under counterfactual manipulation. We unify the three: a
coordinate-indexed scene memory reused for the entire rollout together
with a write-once appearance bank keyed by track identity, retrieved
through banked instance-masked attention at chunk boundaries and upon
re-emergence after occlusion.

\textbf{Reinforcement Learning for Visual Generation.}
RL-based alignment of diffusion and flow models has progressed from
policy-gradient fine-tuning of denoising MDPs (DDPO~\citep{black2023ddpo};
DPOK and Diffusion-DPO for image alignment~\citep{fan2023dpok,wallace2023diffusiondpo};
VADER for video~\citep{prabhudesai2024vader}), through video-scale
preference optimization (SkyReels-V2~\citep{skywork2025skyreels}), to
online group-relative policy optimization for flow matching:
Flow-GRPO introduced ODE-to-SDE exploration~\citep{liu2025flowgrpo},
DanceGRPO extended it to general visual generation~\citep{xue2025dancegrpo},
Phys-AR optimizes physical velocity-error rewards for video
generation~\citep{lin2025reasoning}, and RLVR-World trains world models with
RL~\citep{wu2025rlvrworld}; reward-guided optimization has likewise been
used for identity-preserving image-to-video
generation~\citep{shen2026identity}. On the evaluation side, physical
commonsense benchmarks such as VideoPhy~\citep{bansal2024videophy},
PhysGame, and T2VPhysBench diagnose violations without training, while
VLM judges including Cosmos-Reason1~\citep{nvidia2025cosmosreason} and
VBench-2.0 Motion Rationality establish that vision-language models can
score physical plausibility; driving-specific suites such as
DrivingGen~\citep{zhou2026drivinggen} and WorldLens~\citep{liang2026worldlens}
provide comprehensive but point-estimate evaluations on fixed short
horizons. To our knowledge, we are the first to combine online
flow-matching RL with driving-specific rewards---hierarchical visual
fidelity at scene and instance levels together with a VLM-based physical
plausibility judge---optimized directly on self-generated multi-chunk
rollouts, and the first to evaluate long-horizon driving world models
with quality--horizon decay curves, counterfactual ID-surgery, and
mid-rollout intervention protocols.

\FloatBarrier
\section{Additional Experiments}
\label{app:additional_experiments}

\subsection{Human Evaluation}
\label{app:human_eval}

\textbf{Experimental Setup and Protocol.}
We evaluate human perceptual preference on 60 representative driving scenarios sampled from the test set.
Fifteen independent expert assessors with backgrounds in autonomous driving evaluate WAM-Sim against four baselines: UniMLVG, OpenDWM--CTSD3.5, HorizonDrive, and OmniDreams.
To eliminate bias, all method identities are double-blinded, and video presentation order and spatial placement are randomized.
Assessors perform two complementary tasks: (i) forced-choice pairwise preference voting (Win, Tie, or Loss) for side-by-side video pairs, and (ii) five-point Mean Opinion Score (MOS) rating across four core dimensions: \emph{Visual Fidelity} (perceptual quality and temporal coherence), \emph{Identity Consistency} (long-horizon participant appearance preservation across occlusions), \emph{Physical Realism} (plausibility of agent dynamics and road interactions), and \emph{Location Anchoring} (alignment with spatial and geographic context).

\textbf{Score Aggregation and Inter-Rater Reliability.}
Pairwise preference judgments are aggregated per scenario via majority voting across the 15 assessors to determine scenario-level outcomes.
MOS ratings are averaged per scenario across assessors and subsequently macro-averaged over all 60 scenarios.
Inter-rater reliability across the 15 judges is validated using Fleiss' $\kappa$ on categorical pairwise preferences, demonstrating strong agreement across all baseline comparisons: $\kappa_{\text{UniMLVG}} = 0.927$, $\kappa_{\text{OpenDWM}} = 0.864$, $\kappa_{\text{HorizonDrive}} = 0.939$, and $\kappa_{\text{OmniDreams}} = 0.972$.

\textbf{Quantitative Results and Statistical Analysis.}
As summarized in Table~\ref{tab:human_eval}, WAM-Sim substantially outperforms all baselines in pairwise preference, achieving scenario-level win rates of 98.3\% against UniMLVG ($59/0/1$), 98.3\% against OpenDWM--CTSD3.5 ($59/0/1$), 71.7\% against HorizonDrive ($43/1/16$), and 86.7\% against OmniDreams ($52/0/8$).
Furthermore, WAM-Sim secures the highest MOS across all four perceptual dimensions (Fidelity: 4.30, Identity: 4.40, Physics: 4.50, Location: 4.50).
To assess statistical significance, we conduct two-sided Wilcoxon signed-rank tests ($\alpha=0.05$) on paired scenario-level MOS values (Table~\ref{tab:human_significance}).
WAM-Sim achieves statistically significant gains over UniMLVG, OpenDWM--CTSD3.5, and OmniDreams across all dimensions ($p < 10^{-9}$).
Compared to HorizonDrive (the strongest baseline), WAM-Sim yields statistically significant improvements in Identity ($p=7.65\times10^{-4}$), Physics ($p=4.32\times10^{-8}$), and Location ($p=3.41\times10^{-6}$), confirming that explicit world memory and online RL post-training primarily enhance long-horizon temporal consistency, physical plausibility, and spatial anchoring.

\begin{table}[htbp]
\centering
\caption{
Human evaluation on 60 driving scenarios assessed by 15 experts.
Pairwise preference reports scenario-level win/tie/loss rates against WAM-Sim. MOS is reported on a 1--5 scale.
}
\label{tab:human_eval}
\small
\setlength{\tabcolsep}{4pt}
\renewcommand{\arraystretch}{1.15}
\begin{tabular}{@{}l*{7}{c}@{}}
\toprule
& \multicolumn{3}{c}{\textbf{Pairwise Preference (\%)}}
& \multicolumn{4}{c}{\textbf{Mean Opinion Score}} \\
\cmidrule(lr){2-4}
\cmidrule(lr){5-8}
\textbf{Method}
& \textbf{Win}
& \textbf{Tie}
& \textbf{Loss}
& \textbf{Fidelity}
& \textbf{Identity}
& \textbf{Physics}
& \textbf{Location} \\
\midrule
UniMLVG
& 98.3 & 0.0 & 1.7
& 2.10 & 2.20 & 1.73 & 1.95 \\

OpenDWM--CTSD3.5
& 98.3 & 0.0 & 1.7
& 1.80 & 1.80 & 1.38 & 1.87 \\

HorizonDrive
& 71.7 & 1.7 & 26.7
& 4.10 & 4.00 & 4.00 & 3.90 \\

OmniDreams
& 86.7 & 0.0 & 13.3
& 3.10 & 3.00 & 2.77 & 2.87 \\
\midrule
\textbf{WAM-Sim (Ours)}
& -- & -- & --
& \textbf{4.30}
& \textbf{4.40}
& \textbf{4.50}
& \textbf{4.50} \\
\bottomrule
\end{tabular}
\end{table}

\begin{table}[htbp]
\centering
\caption{
Statistical significance of scenario-level MOS differences.
Two-sided Wilcoxon signed-rank tests compare WAM-Sim with each baseline
using 60 paired scenario-level scores per dimension.
Values below $0.05$ meet the stated significance threshold.
}
\label{tab:human_significance}
\small
\renewcommand{\arraystretch}{1.15}
\begin{tabular}{@{}l*{4}{c}@{}}
\toprule
\textbf{Dimension}
& \textbf{UniMLVG}
& \textbf{OpenDWM--CTSD3.5}
& \textbf{HorizonDrive}
& \textbf{OmniDreams} \\
\midrule
Fidelity
& $7.29\times10^{-12}$
& $7.52\times10^{-12}$
& $0.0546$
& $2.40\times10^{-10}$ \\

Identity
& $7.58\times10^{-12}$
& $7.28\times10^{-12}$
& $7.65\times10^{-4}$
& $2.58\times10^{-10}$ \\

Physics
& $7.52\times10^{-12}$
& $5.96\times10^{-12}$
& $4.32\times10^{-8}$
& $6.49\times10^{-12}$ \\

Location
& $2.78\times10^{-12}$
& $6.25\times10^{-12}$
& $3.41\times10^{-6}$
& $5.17\times10^{-12}$ \\
\bottomrule
\end{tabular}
\end{table}

\subsection{Runtime and Overhead.}
\label{app:runtime}
As detailed in Table~\ref{tab:efficiency_analysis}, WAM-Sim maintains high computational efficiency across resolutions despite incorporating explicit hierarchical memory modules. 
At $704 \times 1280$, WAM-Sim sustains $6.94$~FPS, retaining $86.8\%$ of OmniDreams' throughput ($8.00$~FPS) with a negligible $3.8\%$ increase in peak VRAM ($47.45$~GB vs.\ $45.71$~GB). 
At $256 \times 448$, WAM-Sim reaches $18.63$~FPS, outperforming same-resolution baselines UniMLVG ($1.54$~FPS) and OpenDWM--CTSD3.5 ($1.08$~FPS) by $12.1\times$ and $17.3\times$, respectively. 
Although structural, scene, and instance memory branches expand total parameter capacity from $2.0\text{B}$ to $3.1\text{B}$ ($+55.0\%$), memory retrieval via temporally routed and instance-masked cross-attention incurs minimal latency ($+0.54\text{s}$ per 93-frame chunk). 
\begin{table}[t]
\centering
\caption{Efficiency comparison of driving world models during long-horizon rollouts.}
\label{tab:efficiency_analysis}
\scriptsize
\setlength{\tabcolsep}{5pt}
\renewcommand{\arraystretch}{1.15}

\resizebox{\columnwidth}{!}{%
\begin{tabular}{l|c|c|c|c|c}
\toprule
\textbf{Method}
& \textbf{Resolution}
& \textbf{Params (B)}
& \textbf{Latency (s) $\downarrow$}
& \textbf{Throughput (FPS) $\uparrow$}
& \textbf{VRAM (GB) $\downarrow$} \\
\midrule

Horizondrive         & $384 \times 768$& 1.3 & 3.51 & 4.31 & \textbf{20.11} \\
UniMLVG         & $256 \times 448$& 2.0 & 10.70  & 1.54 & 25.32  \\
OpenDWM-CTSD3.5         & $256 \times 448$& 2.5 & 14.98 & 1.08 & 32.67  \\
Omnidreams         & $704 \times 1280$& 2.0 & 0.65  & 8.00 & 45.71  \\

\midrule

\textbf{Ours}
    & $256 \times 448$& 3.1& \textbf{0.60}  & \textbf{18.63} & 45.21 \\

\textbf{Ours}
    & $704 \times 1280$& 3.1 & 1.19  & 6.94 & 47.45 \\

\bottomrule
\end{tabular}%
}
\end{table}

\subsection{Quality-Horizon Decay Analysis.}
We evaluate temporal stability across $10\,\mathrm{s}$, $30\,\mathrm{s}$, and $60\,\mathrm{s}$ autoregressive rollouts (Fig.~\ref{fig:horizon-ablation}). 
Without world memory, compounding autoregressive errors cause severe degradation from $10\,\mathrm{s}$ to $60\,\mathrm{s}$: 
Subject Coherence drops from $0.535$ to $0.322$, Scene Similarity degrades from $0.791$ to $0.697$, and FVD deteriorates from $197.77$ to $286.54$. 
In contrast, WAM-Sim effectively bounds drift across extended horizons, maintaining high instance coherence ($0.547$), scene similarity ($0.828$), and visual fidelity ($\text{FVD} = 137.92$) at $60\,\mathrm{s}$.

\begin{figure}[tbp]
  \centering
  \includegraphics[width=\linewidth]{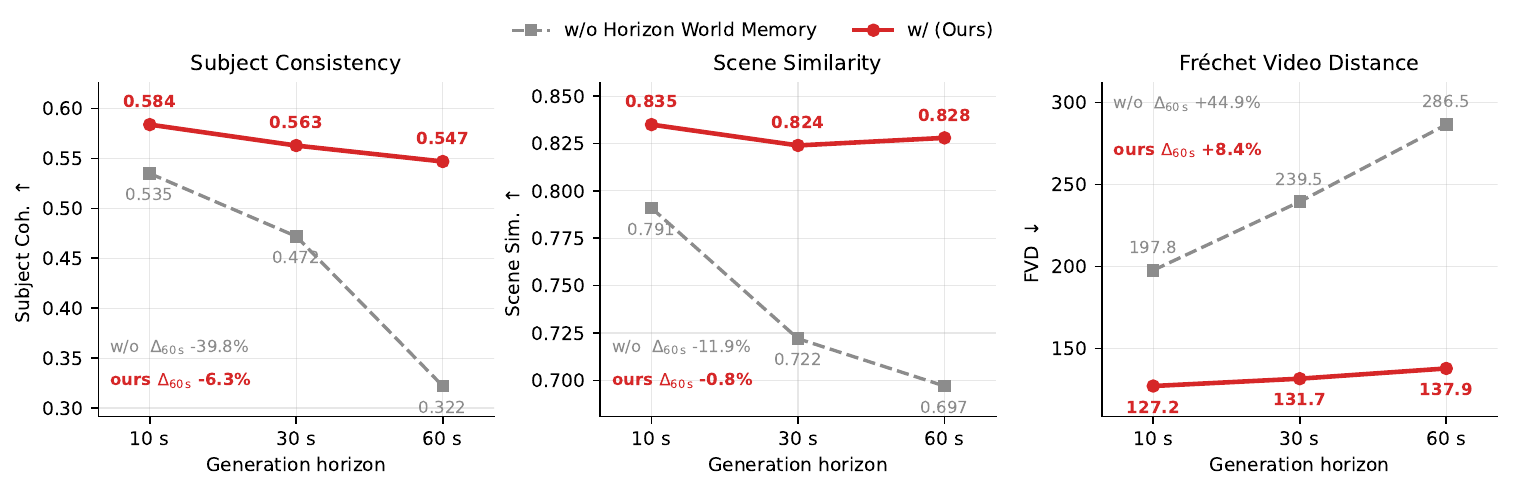}
  \caption{Quality versus generation horizon. World memory reduces quality degradation from 10\,s to 60\,s.}
  \label{fig:horizon-ablation}
\end{figure}

\FloatBarrier

\section{Implementation Details}
\label{app:arch}

We present the architectural specifications, stage-wise training workflow, and reward formulations of WAM-Sim. The framework is trained in four progressive stages: structural control pre-training, scene-memory alignment, instance-memory registration, and online reinforcement post-training. Key model configurations and training parameters are summarized in Tables~\ref{tab:model_config}, \ref{tab:training_config}, and \ref{tab:rl_settings}.

\subsection{Model Configuration}
\label{app:model_config}
WAM-Sim is built on a 28-block 3D Diffusion Transformer (DiT) backbone with a hidden dimension of 2048 and 16 attention heads ($D_{\text{head}}=128$), conditioned on text embeddings from a frozen Qwen2.5-VL-7B encoder. Input videos at $704 \times 1280$ resolution (93 frames at 10~FPS) are encoded by a Wan2.1 Video VAE with $8\times$ spatial and $4\times$ temporal compression into a $24 \times 88 \times 160$ latent grid. Applying $2 \times 2 \times 1$ spatio-temporal patchification yields 84,480 video tokens per chunk, modulated via 3D RoPE and timestep-conditioned adaLN-zero layers. As summarized in Table~\ref{tab:model_config}, Hierarchical World Memory conditions this shared sequence across three scale-specific tiers: (i) \emph{Structural Memory} injects VAE-aligned HD map latents as additive conditioning to establish trajectory-aligned geometry; (ii) \emph{Scene Memory} retrieves one geolocated reference frame ($P$ tokens) per temporal segment, injecting background appearance via segment-masked cross-attention with zero-initialized adaLN gating; and (iii) \emph{Instance Memory} extracts multi-frame appearance crops ($K_f=4$ frames, $K_p=64$ tokens per crop, capping at 32 active instances) into a write-once track-keyed bank, updating foreground tokens via instance-masked attention while bypassing background regions.

\begin{table}[htbp]
    \centering
    \caption{Training settings for Drive DiffusionNFT.}
    \label{tab:rl_settings}
    \small
    \setlength{\tabcolsep}{6pt}
    \renewcommand{\arraystretch}{1.15}
    \begin{tabular}{lc}
        \toprule
        \textbf{Hyperparameter} & \textbf{Value} \\
        \midrule
        Group size $G$ & 16 \\
        Sampling steps & 35 \\
        Guidance scale & 1.0 \\
        Flow shift & 5.0 \\
        NFT mixing coefficient $\beta$ & 1.0 \\
        Training timestep fraction & 0.8 \\
        Reward weights $(\lambda_{\mathrm{scene}},\lambda_{\mathrm{obj}},\lambda_{\mathrm{phy}})$
        & 0.4, 0.3, 0.3 \\
        LoRA rank / alpha & 256 / 256 \\
        Optimizer & AdamW \\
        Learning rate & $1\times10^{-5}$ \\
        Weight decay & $1\times10^{-2}$ \\
        \bottomrule
    \end{tabular}
\end{table}

\begin{table*}[t]
\centering
\caption{Model configuration.
Shared backbone architecture and configurations of the structural, scene,
and instance memory components.}
\label{tab:model_config}
\renewcommand{\arraystretch}{1.15}%
\setlength{\extrarowheight}{2pt}%
\resizebox{\linewidth}{!}{%
\begin{tabular}{@{}lccc@{}}
\toprule
\textbf{Configuration}
& \textbf{Structural Memory}
& \textbf{Scene Memory}
& \textbf{Instance Memory} \\
\midrule

\multicolumn{4}{l}{\textit{Shared Video Representation and Backbone}} \\

Input resolution
& \multicolumn{3}{c}{$704\times1280$} \\

Frames per window
& \multicolumn{3}{c}{$93$} \\

Frame rate
& \multicolumn{3}{c}{$10$ FPS} \\

Video VAE
& \multicolumn{3}{c}{Wan2.1 Video VAE} \\

VAE latent channels
& \multicolumn{3}{c}{$16$} \\

Spatial / temporal compression
& \multicolumn{3}{c}{$8\times$ / $4\times$} \\

Latent shape
& \multicolumn{3}{c}{$24\times88\times160$} \\

Spatial / temporal patch size
& \multicolumn{3}{c}{$2\times2$ / $1$} \\

Video-token grid
& \multicolumn{3}{c}{$24\times44\times80$} \\

Number of video tokens
& \multicolumn{3}{c}{$84{,}480$} \\

Hidden dimension
& \multicolumn{3}{c}{$2048$} \\

Number of DiT blocks
& \multicolumn{3}{c}{$28$} \\

Attention heads
& \multicolumn{3}{c}{$16$} \\

Head dimension
& \multicolumn{3}{c}{$128$} \\

MLP ratio / hidden dimension
& \multicolumn{3}{c}{$4$ / $8192$} \\

Position encoding
& \multicolumn{3}{c}{Learnable 3D RoPE $(t,h,w)$} \\

Normalization
& \multicolumn{3}{c}{Timestep-conditioned adaLN-zero} \\

Text encoder
& \multicolumn{3}{c}{Frozen Qwen2.5-VL-7B (Cosmos-Reason1)} \\

Text embedding dimension
& \multicolumn{3}{c}{$1024$} \\

\midrule
\multicolumn{4}{l}{\textit{World Anchored Memory}} \\

Memory scope
& Trajectory-level
& Temporal-segment-level
& Instance-level \\

Condition source
& HD map
& Geolocated scene reference
& Tracked object appearance \\

Validity domain
& Video-token grid
& Corresponding temporal segment
& Corresponding instance region \\

Representation
& VAE-aligned map latent
& Scene-memory tokens
& Appearance-memory tokens \\

Injection / retrieval
& Additive conditioning
& Segment-masked cross-attn
& Instance-masked attention \\

Persistence
& Trajectory aligned
& Read-only after retrieval
& Persistent across chunks \\

Reference frames
& --
& $1$ per segment
& $K_f=4$ \\

Reference resolution
& --
& \textit{$704\times1280$}
& \textit{$128\times128$} \\

Tokens per reference
& --
& $P$
& $K_p=64$ \\

Tokens per memory entry
& --
& $P$
& $K=256$ \\

Maximum entries
& --
& $N_b$ segments
& $32$ instances \\

Update strategy
& Updated with trajectory
& Fixed after retrieval
& Write once at first appearance \\

Residual modulation
& --
& Zero-initialized adaLN gate
& Zero-initialized adaLN gate \\

\bottomrule
\end{tabular}%
}
\end{table*}

\subsection{Training Configuration}
\label{app:training_config}
WAM-Sim is trained via a progressive four-stage curriculum that incrementally introduces structural guidance, hierarchical memory routing, and online post-training without destabilizing the pretrained backbone. 
Stage~I pre-trains structural control for 130k steps ($\text{LR}=1\times10^{-5}$) using rectified flow to establish trajectory-aligned road geometry. 
Stage~II aligns scene memory for 15k steps ($\text{LR}=3\times10^{-5}$) to inject geolocated background representations via segment-masked cross-attention. 
Stage~III optimizes instance memory for 5k steps ($\text{LR}=3\times10^{-4}$) to register write-once appearance tokens into the track-keyed memory bank. 
Finally, Stage~IV executes online post-training via Drive DiffusionNFT ($\text{LR}=1\times10^{-5}$) with LoRA adapters to align self-generated rollouts against multi-scale visual and physical rewards. 
All stages utilize the AdamW optimizer with constant learning rates, $8\times$ context parallelism, and bfloat16 precision; complete optimization hyperparameters are summarized in Table~\ref{tab:training_config}.

\begin{table}[htbp]
\centering
\caption{Training configuration.
Stage-wise training settings for structural control, scene memory, instance memory,
and reinforcement fine-tuning.}
\label{tab:training_config}
\small
\renewcommand{\arraystretch}{1.15}
\setlength{\tabcolsep}{3pt}

\begin{tabularx}{\linewidth}{@{}>{\raggedright\arraybackslash}p{0.20\linewidth}*{4}{>{\centering\arraybackslash}X}@{}}
\toprule
\textbf{Training Setting}
& Stage I: Structural
& Stage II: Scene
& Stage III: Instance
& Stage IV: Reinforcement \\
\midrule

Training target
& Structural control
& Scene memory
& Instance memory
& Generation quality \\

Initialized from
& Pretrained backbone
& Stage I
& Stage II
& Stage III \\

Trainable modules
& Self-Attn 
& Scene.Cross-attn 
& Instance.Cross-Attn 
& Self-Attn \\

\midrule

Training objective
& Rectified flow
& Rectified flow
& Rectified flow
& DiffusionNFT loss\\

Learning rate
& $1\times10^{-5}$
& $3\times10^{-5}$
& $3\times10^{-4}$
& $1\times10^{-5}$\\

Optimizer
& AdamW
& AdamW
& AdamW
& AdamW \\

Weight decay
& $1\times10^{-3}$
& $1\times10^{-3}$
& $1\times10^{-3}$
& $1\times10^{-2}$\\

LR schedule
& constant
& constant
& constant
& constant\\

Training steps
& 130k
& 15k
& 5k
& 300 \\

Global batch size
& 4
& 4
& 4
& - \\

\midrule

Training resolution
& $704\times1280$
& $704\times1280$
& $704\times1280$
& $704\times1280$ \\

Frames per sample
& $93$
& $93$
& $93$
& $93$ \\

Latent frames
& $24$
& $24$
& $24$
& $24$ \\

Context parallel size
& $8$
& $8$
& $8$
& $8$ \\

Precision
& bf16
& bf16
& bf16
& bf16 \\

Training hardware
& 910b
& 910b
& 910b
& 910b \\

\bottomrule
\end{tabularx}
\end{table}

\subsection{Details of Drive DiffusionNFT Training}
\label{app:rl_details}

Drive DiffusionNFT aligns self-generated rollouts by optimizing a composite objective structured across three hierarchical visual and physical reward components.
First, global visual quality $R_{\mathrm{scene}}$ is computed as the frame-averaged imaging quality score using a SPAQ-pretrained MUSIQ model \citep{huang2024vbench}.
Second, object recognizability $R_{\mathrm{obj}}$ evaluates dynamic traffic participants (pedestrians, cyclists, and vehicles). Subject crops extracted with $1.2\times$ bounding-box expansion are filtered by boundary margin, area, and occlusion, exposure-normalized against ground-truth frames to mitigate luminance gaming, and scored by a frozen ResNet-50 classifier \citep{liang2026worldlens} using dataset-calibrated z-normalized logits.
Third, physical plausibility $R_{\mathrm{phy}}$ is evaluated via SoliReward \citep{lian2026solireward}, an InternVL3-1B model fine-tuned for motion realism and structural deformation assessment under its standard evaluation protocol.

The scalar rewards are group-normalized and linearly aggregated into the overall rollout objective. Optimization parameters, LoRA adapter configurations, and reward weighting factors $(\lambda_{\mathrm{scene}}, \lambda_{\mathrm{obj}}, \lambda_{\mathrm{phy}})$ for online flow-matching post-training are detailed in Table~\ref{tab:rl_settings}.

\FloatBarrier
\section{Dataset Construction}
\label{app:data}

Training a long-horizon driving world model requires more than large-scale video data. The model
must learn temporally consistent visual dynamics together with structured geometric conditions,
persistent object identities, and semantic scene descriptions. We therefore construct a large-scale
in-house autonomous-driving dataset and apply a multi-stage processing pipeline to transform raw
driving logs into structured training samples. This section describes the data sources, construction
pipeline, and final training and evaluation sets.

\subsection{Data Source}
\label{app:data_source}

Our empirical dataset is constructed from a large-scale, in-house autonomous driving corpus gathered across geographically diverse regions in China. 
The raw recordings are organized hierarchically into continuous scenes, each comprising temporally aligned clips with synchronized multi-modal sensor streams and structured annotations. 
Every frame contains a timestamped front-view RGB image ($704\times1280$ at 10~FPS) paired with synchronized vehicle state trajectories, sensor calibration profiles, high-precision LiDAR point clouds, per-frame semantic segmentation masks, and 3D agent bounding boxes parameterized by persistent tracking IDs, positions, and orientations in the ego-vehicle coordinate system. 
Together, these multi-modal streams provide the dense geometric, visual, and instance-level groundings necessary for multi-scale world model training.

\begin{figure}[tbp]
    \centering
    \includegraphics[width=0.92\linewidth]{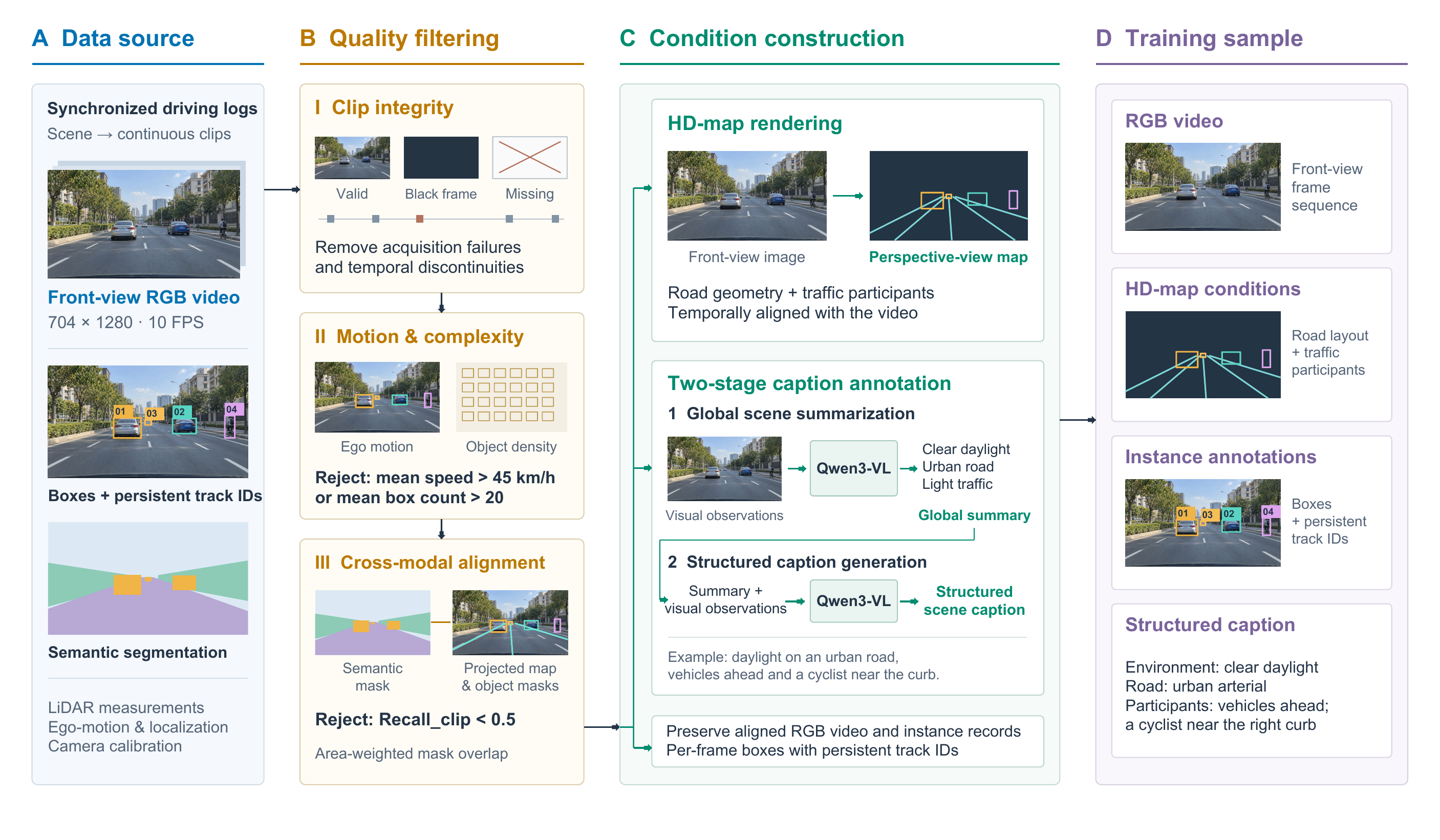}
    \caption{Dataset construction pipeline.}
    \label{fig:data_pipeline}
\end{figure}

\subsection{Data Construction Pipeline}
\label{app:data_pipeline}

Raw driving logs inherently contain sensor anomalies, extreme motion, and misaligned annotations unsuitable for world-model training. To extract high-fidelity, structured training samples, we execute a systematic curation and annotation pipeline (Fig.~\ref{fig:data_pipeline}).

We first enforce data quality through a three-stage filtering protocol. {Stage I: Integrity Filtering} discards sequences exhibiting dropped frames, exposure failures, or temporal discontinuities, preventing the generator from learning low-level sensor artifacts as scene dynamics. {Stage II: Kinematic and Density Filtering} mitigates motion blur and severe occlusion ambiguity by removing clips with average ego-velocities exceeding $45\,\mathrm{km/h}$ or heavily crowded scenes containing more than 20 annotated bounding boxes per frame. {Stage III: Cross-Modal Alignment} ensures precise geometric fidelity between spatial annotations and visual observations. We project HD-map elements and 3D bounding boxes into the 2D image plane to compute their area-weighted overlap ($\mathrm{Recall}_{\mathrm{clip}}$) against corresponding semantic segmentation masks. Sequences with $\mathrm{Recall}_{\mathrm{clip}} < 0.5$ are rejected, systematically eliminating samples afflicted by projection drift or annotation lag.

For the curated sequences, we synthesize aligned conditioning signals to construct the final dataset. {Structural Rendering and Captioning:} We rasterize temporally synchronized HD maps to encode road topology and agent layouts. Concurrently, we generate dense textual descriptions using a two-stage prompting strategy with Qwen3-VL. The vision-language model first extracts global environmental priors (e.g., weather, illumination, traffic density) from the video. This summary subsequently conditions the generation of a comprehensive scene caption detailing fine-grained agent behaviors and ego-state dynamics. An example of the final generated description used as a video-generation prompt is shown in Figure~\ref{fig:generated_prompt_example}.

\begin{figure}[htbp]
    \centering
    \begingroup
    \setlength{\fboxsep}{10pt}
    \setlength{\fboxrule}{0.4pt}
    \fcolorbox{black!18}{black!3}{%
        \begin{minipage}{\dimexpr\linewidth-2\fboxsep-2\fboxrule\relax}
            \small
            \raggedright
            \setlength{\parindent}{0pt}
            \textbf{Example of a Generated Video Prompt}\par
            \vspace{5pt}
            {\color{black!18}\hrule height 0.4pt}
            \vspace{7pt}
            Ego vehicle travels on a multi-lane urban roadway at night under clear conditions with artificial streetlight illumination; wet asphalt surface reflects light, lane markings are visible. Ahead in the center-left lane, a vehicle 10–15m distant drives forward with brake lights illuminated, indicating deceleration. A second vehicle in the center-right lane, 20m ahead, maintains forward motion. On the right sidewalk, 30m ahead, a pedestrian walks parallel to the road. A green traffic light at the upcoming intersection permits progression. A blue directional sign 40m ahead indicates routes to Yan'an Road and Changshu Road; a red vertical sign 15m ahead confirms Yan'an Road.
        \end{minipage}%
    }
    \endgroup
    \caption{Example of a final textual description generated by the two-stage Qwen3-VL annotation pipeline and used as a video-generation prompt.}
    \label{fig:generated_prompt_example}
\end{figure}

Ultimately, each training sample is standardized into four perfectly synchronized modalities: (i) a continuous front-view RGB video; (ii) temporally aligned HD-map sequences; (iii) per-frame 3D instance annotations parameterized by bounding-box coordinates and persistent track identifiers; and (iv) a hierarchical scene caption describing environmental contexts and salient traffic interactions.

\begin{figure}[tbp]
    \centering
    \includegraphics[width=0.92\linewidth]{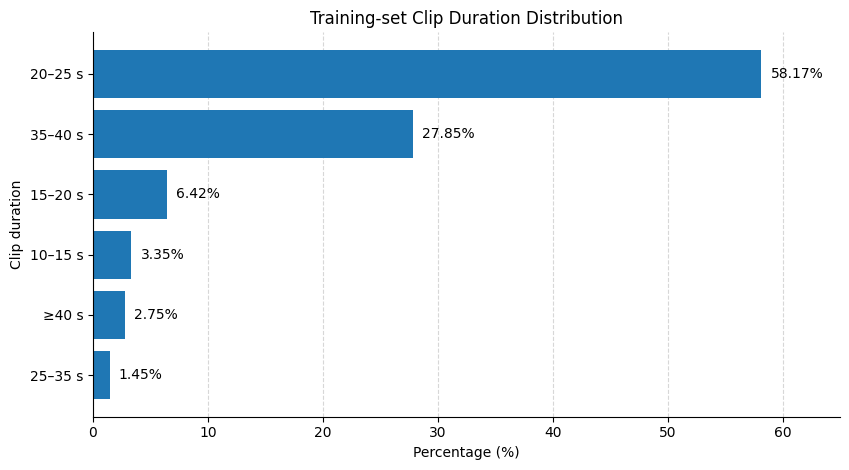}
    \caption{Distribution of clip durations in the training set.}
    \label{fig:duration_distribution}
\end{figure}

\subsection{Training and Evaluation Data}
\label{app:data_split}

The curated training corpus comprises $121,794$ multi-modal driving clips ($\sim$1,000 cumulative hours) sampled at $10\text{ FPS}$ ($704\times1280$ resolution). As illustrated in Fig.~\ref{fig:duration_distribution}, clip durations follow a balanced temporal distribution. The corpus spans diverse geographic regions, urban topology, highway segments, and rural environments across China, encompassing varying illumination (day, dusk, night) and adverse weather conditions (clear, rain, fog). This environmental heterogeneity ensures robust exposure to complex road geometry, lighting transitions, and traffic density variations.

To rigorously assess long-horizon generation and consistency, we establish a standardized evaluation set of 350 curated clips. The evaluation protocol specifically targets challenging edge cases, including severe weather conditions, complex directional maneuvers (e.g., tight turns and unscripted lane changes), and dense interactions with vulnerable road users (pedestrians, cyclists). Furthermore, the evaluation suite isolates identity-sensitive sequences characterized by prolonged tracking occlusions and subsequent re-emergence, providing a stringent benchmark for measuring long-term instance identity persistence and scene stability.

\begin{table}[htbp]
\centering
\caption{Summary of our Self-Collected Dataset.}
\label{tab:dataset_statistics}
\small
\setlength{\tabcolsep}{7pt}
\renewcommand{\arraystretch}{1.15}
\begin{tabular}{l l}
\toprule
\textbf{Statistic} & \textbf{Value} \\
\midrule
Number of training clips & 121,794 \\
Total driving duration & $\sim$1,000 hours \\
Frame resolution & $704\times1280$ \\
Frame rate & 10 FPS \\
Camera view & Front view \\
Geographic coverage & Diverse regions and cities in China \\
Road types & Urban / highway / rural \\
Time of day & Day / dusk / night \\
Weather conditions & Clear / rainy / foggy \\
Structured conditions & HD map + instance annotations + caption \\
Evaluation set & 350 curated clips \\
\bottomrule
\end{tabular}
\end{table}

\subsection{Additional Dataset Details}
\label{app:dataset_details}

Our self-collected corpus comprises driving sequences gathered across diverse urban, highway, and rural environments under varied diurnal and meteorological conditions. To preserve personal privacy, all identifiable elements—including human faces and license plates—undergo automated detection and rigorous spatial blurring prior to processing. Furthermore, location- and identity-associated metadata are anonymized in compliance with standard safety protocols.

The geolocated appearance database, $\mathcal{D}_{\mathrm{geo}}$, indexes reference images using GPS coordinates structured via a spatial k-d tree. Image feature embeddings are pre-computed and cached to eliminate redundant online extraction overhead, restricting spatial query latencies to under $5\,\mathrm{ms}$ during runtime retrieval.

Prior to training, raw logs undergo rigorous multi-stage filtering. Sequences containing sensor dropouts, black frames, or severe temporal discontinuities are systematically discarded. To maintain visual fidelity, we filter out clips featuring excessively high ego-velocities prone to motion blur, alongside over-crowded scenes exceeding dense participant thresholds. Finally, raw sensor streams—including LiDAR point clouds and camera parameters—are projected and rendered into synchronized, model-compatible conditioning representations.
\FloatBarrier

\section{Evaluation and Metric Formulations}
\label{app:eval_protocol}

\subsection{Quality-Horizon Evaluation}
\label{app:quality_horizon}

We evaluate long-horizon generation stability at three rollout durations,
$\mathcal{H}=\{10,30,60\}\ \mathrm{s}$.
Starting from the same initial conditioning, each method is rolled out using the same autoregressive chunk-wise generation protocol. For each target horizon $h\in\mathcal{H}$, we retain the generated sequence from the beginning of the rollout to $h$ seconds and compute all metrics over the corresponding temporal range. The same evaluation samples, temporal horizons, backbone, sampler, and chunk configuration are used for all compared variants.

The ablation isolates the effect of world anchored memory. The baseline retains the structural HD-map conditioning of the backbone but removes the additional scene- and instance-memory readout branches, while our full model enables both. Thus, differences between the two variants reflect the contribution of persistent scene- and instance-level memory under an otherwise identical generation protocol.

We report three complementary measures. \emph{Subject Coherence} evaluates the temporal appearance consistency of foreground traffic participants, while \emph{Scene Similarity} measures the preservation of environment-level visual appearance over the rollout. \emph{Fr\'echet Video Distance} (FVD) measures overall video distributional quality. Higher values are better for Subject Coherence and Scene Similarity, whereas lower values are better for FVD.

For a metric $Q$ and method $m$, let $Q_m(h)$ denote its value evaluated over the rollout up to horizon $h$. We report
\begin{equation}
\mathcal{Q}_m
=
\left\{
Q_m(h)
\mid
h\in\mathcal{H}
\right\},
\label{eq:quality_horizon}
\end{equation}
which characterizes the evolution of generation quality as the autoregressive horizon increases. All methods are evaluated independently at the same three horizons using an identical protocol.

\subsection{Scene Preservation Metrics}
\label{app:scene_metrics}
\textbf{Scene Similarity (SS).}
Scene Similarity measures the visual consistency between the generated video
and the ground-truth video within the scene regions.
Let $\hat{\mathbf{I}}_t$ and $\mathbf{I}^{\mathrm{GT}}_t$ denote the generated
and ground-truth frames at time $t$, respectively, and let
$\mathbf{M}_t \in \{0,1\}^{H \times W}$ denote the corresponding scene mask.
A frozen DINO image encoder $\phi$ extracts patch-level features, and the scene
mask is average-pooled to the DINO patch grid to obtain patch weights
$m_{t,p} \in [0,1]$.

The Scene Similarity of the entire video is defined as
\begin{equation}
\mathrm{SS}
=
\frac{
    \sum_{t=1}^{T}\sum_{p=1}^{P}
    m_{t,p}\,
    \cos\left(
        \phi_p(\hat{\mathbf{I}}_t),
        \phi_p(\mathbf{I}^{\mathrm{GT}}_t)
    \right)
}{
    \sum_{t=1}^{T}\sum_{p=1}^{P}m_{t,p}
},
\end{equation}
where $T$ is the number of video frames, $P$ is the number of DINO patch
tokens, and $\phi_p(\cdot)$ denotes the DINO feature of patch $p$.
A higher SS indicates stronger scene-level appearance and structural
consistency with the ground-truth video.

\subsection{Identity Preservation Metrics}
\label{app:identity_metrics}

We evaluate instance identity preservation using appearance embeddings extracted from generated-video crops. For each visible instance $i$ at frame $t$, we square-expand its ground-truth (GT) bounding box by 10\%, resize the resulting crop to $224\times224$, and extract an appearance embedding using a frozen DINOv2 ViT-B/14 model. The embedding is $\ell_2$-normalized and denoted as $e_{t,i}$. All reported similarities are cosine similarities, and higher values indicate better identity preservation. We use a temporal stride of $S=92$ frames, corresponding to the autoregressive chunk interval.

A track is considered valid for SC@$kS$ if it contains at least one pair of frames separated by $kS$ for which the instance remains continuously visible throughout the entire interval. For occlusion-based metrics, an occlusion event is defined as a maximal contiguous run of invisible frames that is strictly bounded by visible segments of the same track. The pre- and post-occlusion appearance prototypes are constructed only from the contiguous visible segments immediately adjacent to the corresponding occlusion event.

\textbf{Long-range coherence (SC@$kS$).}
For each track, we compare appearance embeddings separated by a temporal gap of
\[
\Delta = kS, \qquad k\in\{1,2,3\},
\]
while requiring the instance to remain continuously visible throughout the interval.

Let $\mathcal{V}_{i,k}$ denote the set of valid starting frames for track $i$:
\[
\mathcal{V}_{i,k}
=
\left\{
t \,\middle|\,
\text{instance } i \text{ remains visible from } t
\text{ through } t+kS
\right\},
\]
and let
\[
\mathcal{T}_k
=
\left\{
i \,\middle|\,
|\mathcal{V}_{i,k}|>0
\right\}
\]
denote the set of valid tracks. We compute
\begin{equation}
\mathrm{SC@}kS
=
\frac{1}{|\mathcal{T}_k|}
\sum_{i\in\mathcal{T}_k}
\frac{1}{|\mathcal{V}_{i,k}|}
\sum_{t\in\mathcal{V}_{i,k}}
e_{t,i}^{\top}e_{t+kS,i},
\qquad
k\in\{1,2,3\}.
\label{eq:sc_metric}
\end{equation}
This metric measures long-range appearance coherence while excluding appearance changes associated with disappearance and re-emergence.

\textbf{Occlusion and re-emergence.}
An occlusion event $u$ is a maximal contiguous interval during which an instance is invisible and is preceded and followed by visible segments of the same track. Let $i(u)$ denote the corresponding instance.

For each event, $\mathcal{P}_u$ contains up to the last $S$ frames from the contiguous visible segment immediately preceding the occlusion, and $\mathcal{Q}_u$ contains up to the first $S$ frames from the contiguous visible segment immediately following re-emergence. We construct appearance prototypes on the two sides of the occlusion as
\begin{equation}
p_u^{-}
=
\operatorname{Norm}
\left(
\frac{1}{|\mathcal{P}_u|}
\sum_{t\in\mathcal{P}_u}
e_{t,i(u)}
\right),
\qquad
q_u^{+}
=
\operatorname{Norm}
\left(
\frac{1}{|\mathcal{Q}_u|}
\sum_{t\in\mathcal{Q}_u}
e_{t,i(u)}
\right),
\label{eq:occ_embedding}
\end{equation}
where $\operatorname{Norm}(\cdot)$ denotes $\ell_2$ normalization.

The re-emergence similarity for event $u$ is then defined as
\begin{equation}
s_u
=
(p_u^{-})^{\top}q_u^{+}.
\label{eq:reemerge_similarity_event}
\end{equation}

\textbf{Re-emergence similarity.}
We report a video-level macro-average to prevent videos containing many occlusion events from dominating the metric. Let $\mathcal{D}$ denote the set of videos containing at least one valid occlusion event, and let $\mathcal{U}_v$ denote the set of valid occlusion events in video $v$. We compute
\begin{equation}
\mathrm{Re\mbox{-}emerg.\ Sim.}
=
\frac{1}{|\mathcal{D}|}
\sum_{v\in\mathcal{D}}
\frac{1}{|\mathcal{U}_v|}
\sum_{u\in\mathcal{U}_v}
s_u.
\label{eq:reemerge_sim}
\end{equation}

\textbf{Re-emergence pass rate.}
We additionally report the fraction of valid occlusion events whose re-emergence similarity exceeds a GT-calibrated threshold $\tau$. Let
\[
\mathcal{U}
=
\bigcup_{v\in\mathcal{D}}
\mathcal{U}_v
\]
denote the set of all valid occlusion events. The pass rate is
\begin{equation}
\mathrm{PassRate}
=
\frac{1}{|\mathcal{U}|}
\sum_{u\in\mathcal{U}}
\mathbf{1}[s_u\geq\tau].
\label{eq:pass_rate}
\end{equation}
We set $\tau=0.803$, corresponding to the median GT re-emergence similarity over all valid occlusion events in the evaluation set. The threshold is determined once from the GT videos and is fixed for all evaluated methods.

\textbf{Occ-ID@$mS$.}
To analyze identity preservation under different occlusion durations, we group events according to their occlusion length $L_u$. For the $m$-th duration interval, we define
\begin{equation}
\mathcal{B}_m
=
\left\{
u \,\middle|\,
(m-1)S<L_u\leq mS
\right\},
\label{eq:occ_bin}
\end{equation}
and compute
\begin{equation}
\mathrm{Occ\mbox{-}ID@}mS
=
\frac{1}{|\mathcal{B}_m|}
\sum_{u\in\mathcal{B}_m}
s_u.
\label{eq:occ_id}
\end{equation}

In our test set, we report Occ-ID@1S and Occ-ID@2S, corresponding to occlusion lengths of $(0,92]$ and $(92,184]$ frames, respectively. The Occ-ID@2S interval contains only three valid events, and no valid event has an occlusion duration longer than 184 frames.

Re-emergence similarity is macro-averaged at the video level, whereas Pass Rate and Occ-ID are averaged over individual occlusion events.

\textbf{Average score.}
The reported Avg. score is the arithmetic mean of the seven metrics in Table~\ref{tab:instance_identity}: SC@1S, SC@2S, SC@3S, Re-emergence Similarity, Re-emergence Pass Rate, Occ-ID@1S, and Occ-ID@2S.

The GT videos are used only to determine track identities, visibility, occlusion events, bounding boxes, and the calibration threshold $\tau$. All appearance similarities reported for each evaluated method are computed exclusively from the generated videos.

\subsection{RL Evaluation Metrics}
\label{app:rl_eval_metrics}
\textbf{Subject Fidelity.}
We measure whether the conditioned dynamic subjects are detectable in the generated video with a GT-gated detector protocol. Conditioning boxes (per-frame tight AABBs of the conditioned pedestrians, cyclists, cars and vehicles) are taken from the ground-truth annotations. A YOLO26s detector trained on BDD100k is run on every frame of the generated video; each conditioning box is greedily matched to the highest-IoU same-class detection. To exclude annotation and visibility ambiguity from the denominator, the identical detection-and-matching procedure is run on the ground-truth video, and only instances the detector recovers in the GT frames count as valid.

\textbf{Physical Plausibility.}
We use Qwen3.5-9B as a no-reference VLM judge to evaluate pedestrian motion plausibility. Each video is evaluated with two binary questions covering motion anatomy and limb integrity. We use the normalized probability of ``yes'' at the first generated token rather than a sampled binary response. Specifically, the probability is computed by renormalizing the logits over the predefined yes/no token variants. The final score is the mean $P(\mathrm{yes})$ across the two questions. Only videos with clearly visible pedestrians are included in this evaluation.

\begin{tcolorbox}[
    colback=gray!4,
    colframe=gray!40,
    boxrule=0.5pt,
    arc=2pt,
    left=6pt,
    right=6pt,
    top=5pt,
    bottom=5pt,
    title=\textbf{Physical-Plausibility Prompts},
    fonttitle=\small,
    coltitle=black
]
\small

\textbf{Motion Anatomy:}
Do all walking or running pedestrians show perfectly natural, anatomically plausible limb movement throughout the entire video, with zero unnatural joint motion and zero abnormal limb trajectories?
\textit{(yes or no)}

\vspace{0.5em}

\textbf{Limb Integrity:}
For moving pedestrians, do their arms and legs remain clearly distinguishable throughout the motion, without limbs merging into the body, merging with each other, disappearing, or becoming visually entangled?
\textit{(yes or no)}

\end{tcolorbox}

\FloatBarrier
\section{Limitations and Future Work}

Despite its strong long-horizon consistency, WAM-Sim exhibits three primary limitations. 
First, memory retrieval latency and attention memory footprint scale linearly with the number of tracked foreground instances, 
introducing computational overhead in hyper-dense traffic scenes containing over 100 concurrent agents. 
Second, the write-once instance appearance bank assumes rigid visual properties; 
consequently, extreme non-rigid state transitions (e.g., severe vehicle damage during collisions or drastic open/close deformations of commercial truck beds) are constrained by the initial-observation tokens. 
Third, background scene retrieval depends on geolocated reference availability, which may degrade to ungrounded autoregressive generation in unmapped or highly dynamic construction zones.

To address these challenges, future iterations will pursue three key research directions:
1) Integrating explicit 3D visual representations, such as online-updated Gaussian Splatting or neural radiance fields, into the memory substrate to enable unified multi-view spatial consistency and continuous geometry-aware retrieval under counterfactual maneuvers.
2) Formulating a parametric write-update memory mechanism that selectively incorporates visual state changes across significant illumination or topological shifts while preserving fundamental instance identity invariants.
3) Developing dynamic instance-grouping and memory-compression protocols to bound memory complexity in extreme ultra-dense urban traffic scenarios.
\FloatBarrier
\section{Ethics, Broader Impact, and Reproducibility Assurance}
\label{app:ethics_impact}

WAM-Sim advances autonomous driving safety by enabling controllable, zero-cost edge-case simulation over extended horizons. While high-fidelity video generation presents potential dual-use risks regarding synthetic media misuse, these concerns can be effectively mitigated through digital watermarking protocols and restricted access frameworks for generative models.

To ensure full reproducibility, we release our complete training and evaluation code, model checkpoints, and dataset processing pipelines. Detailed architectural configurations, optimization hyperparameters, and hardware environments are thoroughly documented in Appendices~\ref{app:arch} and~\ref{app:data}.

\FloatBarrier
\endgroup

\end{document}